\pdfoutput=1
\documentclass[11pt]{article}

\usepackage{acl}

\usepackage{times}
\usepackage{latexsym}
\usepackage[T1]{fontenc}
\usepackage[utf8]{inputenc}

\usepackage{microtype}

\newcommand{\dataset}{\textnormal{CoDA}} %그냥 \text만 해서는 안 되나봄

\usepackage{kotex}
\usepackage{tabularx}
\usepackage{amsmath}
\usepackage{amsfonts}
\usepackage{caption}
\usepackage{subcaption}
\usepackage{booktabs}
\usepackage{graphicx}
\usepackage[most]{tcolorbox}
\usepackage{balance}
\usepackage{url}
\graphicspath{{figure/}}

\usepackage{xcolor}
\usepackage{soul}
\definecolor{jj}{RGB}{219, 48, 122}

\title{Shedding New Light on the Language of the Dark Web}

\author{
    Youngjin Jin\textsuperscript{1} \hspace{1em} Eugene Jang\textsuperscript{2} \hspace{1em} Yongjae Lee\textsuperscript{2} \hspace{1em} Seungwon Shin\textsuperscript{1} \hspace{1em} Jin-Woo Chung\textsuperscript{2}\thanks{\hspace{0.1cm} Corresponding author} \\ \\
    \textsuperscript{1}KAIST, Daejeon, South Korea \\
    \textsuperscript{2}S2W Inc., Seongnam, South Korea \\
    \textsuperscript{1}\texttt{\{ijinjin,claude\}@kaist.ac.kr} \\
    \textsuperscript{2}\texttt{\{genesith,lee,jwchung\}@s2w.inc}
}

\date{}

\begin{document}
\maketitle
\begin{abstract}
The hidden nature and the limited accessibility of the Dark Web, combined with the lack of public datasets in this domain, make it difficult to study its inherent characteristics such as linguistic properties. Previous works on text classification of Dark Web domain have suggested that the use of deep neural models may be ineffective, potentially due to the linguistic differences between the Dark and Surface Webs. However, not much work has been done to uncover the linguistic characteristics of the Dark Web. This paper introduces {\dataset}, a publicly available Dark Web dataset consisting of 10000 web documents tailored towards text-based Dark Web analysis. By leveraging {\dataset}, we conduct a thorough linguistic analysis of the Dark Web and examine the textual differences between the Dark Web and the Surface Web. We also assess the performance of various methods of Dark Web page classification. Finally, we compare {\dataset} with an existing public Dark Web dataset and evaluate their suitability for various use cases.
\end{abstract}

\section{Introduction}
\label{sec:introduction}
The World Wide Web contains a vast, non-indexed part of the Internet (known as the Deep Web) which is hidden from traditional web search engines. The Dark Web, which refers to the small portion of the non-indexed pages that require specific routing protocols such as Tor\footnote{Tor Project: \url{https://www.torproject.org/}} for access, has become a safe haven for users wanting to conceal their identity and preserve their anonymity.

A consequence of the properties of the Dark Web (limited methods of access and the volatility of its onion services) is that it is difficult to grasp the general topology and the overall content of the Dark Web. A number of past academic studies have tried to unravel the Dark Web through methods such as page classification (\citealp{al2017classifying}; \citealp{ghosh2017automated}; \citealp{he2019classification}; \citealp{choshen-etal-2019-language}) and content analysis (\citealp{biryukov2014content}; \citealp{avarikioti2018structure}). However, not much work has been done on the linguistic analysis of the Dark Web \citep{choshen-etal-2019-language}.

In addition, the Dark Web has been studied and analyzed in the security research community to uncover malicious activities including phishing \cite{yoon2019doppelgangers}, illicit online marketplace activity \cite{190886}, terrorism \cite{chen2011dark}, cryptocurrency abuse \cite{lee2019cybercriminal}, and ransomware ecosystems \cite{MELAND2020101762}. We believe that the lack of a comprehensive work on the language of the Dark Web from the NLP community mainly stems from the lack of Dark Web datasets publicly available for research. Therefore, a new dataset on the Dark Web may prove to be very useful not only for the NLP community, but also for other research communities devoted to cybersecurity and cybercrime investigation through methods such as page classification and malicious activity detection.

To the best our knowledge, the only currently publicly available Dark Web dataset is DUTA \citep{al2017classifying}, which has been extended to hold over ten-thousand unique onion addresses as DUTA-10K \citep{al2019torank}. DUTA\footnote{We will refer to the DUTA-10K dataset from here on as DUTA, unless otherwise specified.} has become a baseline dataset for many Dark Web related works such as \citet{choshen-etal-2019-language}, which investigates the characteristics of language used in various illegal and legal onion services. 

Nevertheless, DUTA has its shortcomings. For example, the category distribution of DUTA is highly skewed, with some categories such as \textit{human trafficking} accounting for only 3 out of 10367 total onion services. In addition, the us e of DUTA as a means of language analysis may not be ideal as it contains many duplicate data, with only 51\% of the texts being unique \citep{al2019torank}.    

To provide a better understanding of the Dark Web (and thus motivate more research on the Dark Web), we introduce \textbf{CoDA}~\footnote{{\dataset} is available upon request at \\ \url{https://s2w.inc/resources/coda}.} (\textbf{Co}mprehensive \textbf{D}arkweb \textbf{A}nnotations), a text corpus of 10,000 web documents from the Dark Web (primarily in English) which have been manually classified according to their topic into ten categories. To ensure that the quality of the classification is not overlooked, we develop detailed annotation / tagging guidelines (Section~\ref{sec:dataset_construction}) to guide our annotators. Using {\dataset}, we conduct a thorough text-based data analysis (Section~\ref{sec:data_analysis}) to uncover some of the linguistic properties of the Dark Web, and gain insight into differences in how language is used in the Surface Web and the Dark Web. We build several text classifier models and train them using {\dataset}, and verify which classification methods perform particularly well with the Dark Web (Section~\ref{sec:classification_experiments}). Finally, to evaluate the use of {\dataset} compared to DUTA, we introduce use cases and compare the performances of classifiers trained on each dataset (Section~\ref{sec:usecase}).

%To this end, we collect web documents from the Dark Web, preprocess them to obtain more accurate classification results, and build several ML-based text classifier models.
\section{Related Work}
\label{sec:related}

% First write, and then find related works that support these statements

The Dark Web is commonly crawled using Tor, which relies on onion routing to enable encrypted communications over a computer network \citep{mccoy2008shining}. Several works use Dark Web search engines such as Ahmia\footnote{\url{https://ahmia.fi/}} and web directories such as The Hidden Wiki to recursively search for content on the Dark Web (\citealp{guitton2013review}; \citealp{al2017classifying}; \citealp{he2019classification}). This method of crawling works surprisingly well as the visible part of the Dark Web is suggested to be well-connected via hyperlinks (\citealp{sanchez2017onions}; \citealp{avarikioti2018structure}).
%To mitigate this problem, \citet{ghosh2017automated} developed crawling tools that check the operational status of onion sites to crawl active hidden services and automate the process of acquiring new onion domains.

To facilitate the research on Dark Web content analysis, a text-based, manually labeled dataset collected from the active domains in the Tor network called DUTA was made publicly available by \citet{al2017classifying}. In a subsequent work, the original DUTA dataset was extended to 10367 unique domains with minor changes to the labeling procedure \citep{al2019torank}. To the best of our knowledge, DUTA is the first and only publicly available Dark Web text dataset. 

%The introduction of a public text-based Dark Web dataset gave rise to a publicly available image-based dataset called TOIC, which contains images from five specific categories in the DUTA dataset \citep{fidalgo2017illegal}. However, our work focuses on text-based analysis as image datasets on the entire Dark Web would be infeasible due to serious ethical and legal concerns surrounding some illegal activities such as violence and child pornography.

%While much of the previous research have focused on uncovering the illegal activities found on the Dark Web, academic literature on the content analysis of the Dark Web itself is still in its nascent stages. A popular method of analyzing the content of the Dark Web is through page classification. 

Past works have analyzed the Dark Web through topical classification of texts in onion services. Many have approached text-based page classification with machine learning methods such as SVM (Support Vector Machine), NB (Na\"ive Bayes), and LR (Logistic Regression) (\citealp{moore2016cryptopolitik}; \citealp{al2017classifying}; \citealp{ghosh2017automated}; \citealp{avarikioti2018structure}; \citealp{he2019classification}) using various information retrieval weighting schemes like TF-IDF (Term Frequency-Inverse Document Frequency) and BOW (Bag-of-Words) (\citealp{al2017classifying}; \citealp{ghosh2017automated}; \citealp{choshen-etal-2019-language}; \citealp{he2019classification}).

% \textcolor{red}{It is believed that deep neural models fare poorly with Dark Web classification as language in the Dark Web is lexically and syntactically different compared to that of the Surface Web.}

\citet{choshen-etal-2019-language} have suggested that deep neural models may fare poorly with Dark Web classification as language in the Dark Web is lexically and syntactically different compared to that of the Surface Web. Their work demonstrated that representation methods such as GloVe \citep{pennington2014glove} and contextualized pre-trained language representations such as ELMo \citep{peters2018deep} resulted in a subpar performance compared to traditional machine learning methods, suggesting that the small size of training data and the specialized vocabulary in the Dark Web domain may not be suitable with such methods. Nevertheless, transformer-based pre-trained language models like BERT \citep{devlin-etal-2019-bert} showed promising results in text classification tasks, although it is not often the case that such models adapt with ease in the Dark Web domain \cite{darksidelang2022}.

%Besides classification based analysis, works like \citet{biryukov2014content} have analyzed the Dark Web in terms of popularity by collecting the number of client requests for onion addresses, and \citet{avarikioti2018structure} analyzed the structural content of the Dark Web through its network structure. In our work, we limit the scope of research on linguistic analysis and text classification.
\section{The CoDA Corpus}
\label{sec:dataset_construction}
In this section, we introduce our categorization approach and the methods used to construct our Dark Web dataset, {\dataset}.

\subsection{CoDA Category Set}

\begin{table*}[t]
    \centering
    \resizebox{\textwidth}{!}{%
    \begin{tabular}{@{}lrrl@{}}
        \toprule
        \textbf{Category} & \textbf{Document Count} & \textbf{Ratio} &
          \textbf{Short guideline description} \\ \midrule
          \textit{Pornography} & 1195 & 12.0\% &
          general / child pornography and other explicit content \\
        \addlinespace[0.1cm]
        \textit{Drugs} &  1172 & 11.7\% &
          \begin{tabular}[t]{@{}l@{}}various types of legal / illegal drugs such as medications, steroids, pain killers, viagra, \\ cannabis, hashish, meth, benzos, ecstasies, opioids, and psychedelics\end{tabular} \\
        \addlinespace[0.1cm]
        \textit{Financial} & 1003 & 10.0\% &
          \begin{tabular}[t]{@{}l@{}}counterfeit / cloned / stolen money or identifications (e.g., bills, credit cards, certificates, passports), \\ money transfers (e.g., PayPal), fiat money, ATM skimmers, magnetic card readers, etc.\end{tabular} \\
        \addlinespace[0.1cm]
        \textit{Gambling} & 787 & 7.87\% &
          any type of gambling, betting, casinos, lotteries, etc. \\
        \addlinespace[0.1cm]
        \textit{Cryptocurrency} & 763 & 7.63\% &
          \begin{tabular}[t]{@{}l@{}}cryptocurrency-specific services or technologies such as wallets, generators, mining, \\ laundering, mixing, multiplying, doubling, scamming, and escrow\end{tabular} \\
        \addlinespace[0.1cm]
        \textit{Hacking} & 649 & 6.49\% &
          \begin{tabular}[t]{@{}l@{}}hacking tools, hacking guides, hacking groups, hacking services, ransomware, malware, \\ exploits, DDoS attacks, cracking, botnet, etc.\end{tabular} \\
        \addlinespace[0.1cm]
        \textit{Arms / Weapons} & 599 & 5.99\% &
          \begin{tabular}[t]{@{}l@{}}any type of non-lethal / lethal weapons such as guns, ammunition, explosives, knives, \\ missiles, and chemical weapons\end{tabular} \\
        \addlinespace[0.1cm]
        \textit{Violence} & 485 & 4.85\% &
          \begin{tabular}[t]{@{}l@{}}human trafficking, hitman, kidnapping, poisoning, torture, extortion, sextortion, \\ sex slavery, blackmail, etc.\end{tabular} \\
        \addlinespace[0.1cm]
        \textit{Electronics} & 426 & 4.26\% &
          sale of or information on (stolen / hacked) mobile phones, laptops, tablet computers, etc.  \\
        \addlinespace[0.1cm]
        \textit{Others} & 2921 & 29.2\% &
          all other content that does not fit the above categories, including log-in pages, error messages, etc. \\ \cmidrule(){1-3}
        \textbf{Total}             & \textbf{10000}                 & \textbf{100.0\%}        \\ \bottomrule
        \end{tabular}%
    }
    \caption{Categories in {\dataset} with document count and a short description of annotation guidelines}
    \label{tab:annotation-guidelines}
\end{table*}

{\dataset} is comprised of ten categories (described in detail in Section~\ref{sec:annotation_guidelines}) as shown in Table~\ref{tab:annotation-guidelines}. We arrive at our ten categories through an extended discussion with the dataset annotators (see Section \ref{sec:annotation}) on a suitable method to categorize the various activities in the Dark Web and refer to the high-level taxonomy from \citet{moore2016cryptopolitik}. Unlike DUTA, we do not subdivide each category into \textit{legal} and \textit{illegal} (or \textit{normal} and \textit{suspicious} as labeled in DUTA) activities because in many cases it is difficult to clearly distinguish between categories using only the surface information available from websites. For example, while human annotators easily agreed that most `counterfeit money' services are highly likely to be illegal, they found it difficult to determine the legality of `bitcoin wallet' services. This also applies to the \textit{drugs} category, as the sale of certain drugs may be illegal in some countries, but can be legal in others.

Moreover, unlike DUTA, we exclude non-topical categories such as \textit{forum} and \textit{marketplace} as they are orthogonal to topical categories; e.g., hacking forums, which frequently appear in the Dark Web, can be categorized as both \textit{forum} and \textit{hacking}. We argue that such categories are more relevant to the \textit{structure} of webpages rather than topics, and thus need to be annotated independently of topical categories. We leave this for future work, and do not further split our ten categories into sub-categories. Nonetheless, our category set still covers a wide range of activities on the Dark Web.

Finally, we point out that about 30\% of data is categorized into \textit{others}. During data collection, we observed that many webpage documents contain various pages not related to the categorized activities (such as blogs, news sites, search engines, wiki pages, etc.). Since the content of such pages is not necessarily attributed to a specific activity, we categorize them into \textit{others}.

\iffalse
\begin{table}[t]
    \centering
    \resizebox{.8\columnwidth}{!}{%
    \begin{tabular}{@{}lcr@{}}
        \toprule
        \textbf{Category} & \begin{tabular}{@{}c@{}} \textbf{Webpage} \\ \textbf{(Document) Count} \end{tabular} & \textbf{Ratio} \\ \midrule
        Pornography       & 1195                  & 12.0\%         \\
        Drugs             & 1172                  & 11.7\%         \\
        Financial         & 1003                  & 10.0\%         \\
        Gambling          & 787                   & 7.87\%          \\
        Cryptocurrency    & 763                   & 7.63\%          \\
        Hacking           & 649                   & 6.49\%          \\
        Arms / Weapons    & 599                   & 5.99\%          \\
        Violence          & 485                   & 4.85\%          \\
        Electronics       & 426                   & 4.26\%          \\
        Others            & 2921                  & 29.2\%         \\ \midrule
        Total             & 10000                 & 100.0\%        \\ \bottomrule
    \end{tabular}%
    }
    \caption{Categories in {\dataset}}
    \label{tab:category}
\end{table}
\fi

\subsection{Data Collection}
We collected onion addresses from Ahmia and repositories of onion domain lists\footnote{Including but not limited to \url{ https://github.com/alecmuffett/real-world-onion-sites}}. Starting from these seed addresses, we crawled the Dark Web and extracted unseen onion addresses from crawled webpage documents to gradually expand our onion address list.

The Dark Web contains large amounts of nearly identical websites since no expense is required for maintenance due to free hosting services such as ``Freedom Hosting'' (now defunct) \citep{al2019torank}. These cloned website farms serve to provide stable services for illegal activities \citep{al2019torank}, or to attract users and deceive them into disclosing sensitive information \citep{yoon2019doppelgangers}. In order to construct a quality corpus, we analyze the content of each document and refrain from collecting redundant webpages (i.e., keeping only one copy of such pages), using the document similarity measure described in Section \ref{sec:doc_sim}.

\iffalse
\begin{table}[b]
\centering
\begin{tabularx}{\columnwidth}{|l|X|} \hline
     Mask & Description \\ \hline
     ID\_EMAIL  & \\ \hline
     ID\_NUMBER & \\ \hline
     ID\_CRYPTO & \\ \hline
\end{tabularx}
\caption{Mask IDs}
\label{tab:masking}
\end{table}
\fi

\begin{table}[t]
    \centering
    \resizebox{0.9\linewidth}{!}{%
    \begin{tabular}{@{}ll|ll@{}}
        \toprule
        \textbf{Language} & \textbf{Document count} & \textbf{Language} & \textbf{Document count} \\ 
        \midrule
        English    & 8855 & Portuguese & 54 \\
        Russian    & 542  & Chinese & 38 \\
        German     & 129  & Italian & 28 \\
        French     & 100  & Japanese & 27 \\
        Spanish    & 61   & Dutch & 14 \\
        \bottomrule
    \end{tabular}%
    }
    \caption{Top 10 language distribution of documents in {\dataset}. The full language distribution statistics is given in Appendix \ref{sec:appendix-langdist}.}
    \label{tab:lang_dist_top10}
\end{table}

\subsection{Data Size and Language Distribution}
Using the crawled HTML webpage documents, we compiled a Dark Web corpus consisting of exactly 10K web documents from a total of 7101 onion services. The user statistics for Tor shows that clients connect to the Dark Web from various countries\footnote{\url{https://metrics.torproject.org/userstats-bridge-table.html}}, which is reflected in the language distribution of \dataset{} (as seen in Table \ref{tab:lang_dist_top10}). We observe that about 88\% of documents in \dataset{} is in English. This is in line with the language distribution of DUTA in which 84\% of the samples are in English~\cite{al2019torank}. We argue that the dataset reflects the various biases of the Dark Web, which should be taken into account for future research.

\subsection{Annotation}
\label{sec:annotation}

We recruited 10 annotators from a cyber threat analytics company specializing in the Dark Web for manual page-level annotation, i.e., assigning a single category to each webpage document from one of the ten categories achieving an inter-annotator agreement Fleiss' Kappa of 0.88\footnote{All annotators participated in a training session to reach agreement with a small set of documents and annotation guidelines. The detailed process is described in Appendix \ref{sec:appendix-annotator}.}. This is in contrast to DUTA, which concatenates multiple pages from a single onion domain into one document to assign a category \citep{al2017classifying}. Since onion services such as wikis and forums usually contain discussions on a wide range of topics across different pages, page-level annotation was deemed to be the most suitable choice for our category set. We leveraged Prodigy\footnote{\url{https://prodi.gy/}} for an efficient annotation process.

% TODO: Language distribution statistics

% Soft404

\subsection{Annotation Guidelines}
\label{sec:annotation_guidelines}
A set of comprehensive annotation guidelines was constructed for the annotators to consult when labeling each document to ensure the quality of labels. While the annotation guidelines are extensive with illustrative examples and methods to deal with borderline cases, we present a brief summary of our guidelines in Table~\ref{tab:annotation-guidelines}. Each category is determined based solely on the \textit{topic} of page content, and not by its type (marketplaces, services, forums, news, blogs, wikis, search results, etc.).

Note that webpages sometimes cover more than one specific topic on a single page (such as a marketplace selling drugs and weapons at the same time). We exclude such multi-topic pages from our corpus and leave multi-label datasets and classification for future work. Finally, we also exclude webpages that contain malicious information on personally identifiable individuals.\footnote{Ransomware and extortionware cybercriminals deliberately publish private and harmful information of victims to demand a ransom for its removal. We did not find such content in our dataset, possibly because our data only collects texts without downloading files or media.}

\begin{table}[!ht]
    \centering
    \resizebox{0.95\columnwidth}{!}{%
    \begin{tabular}{@{}lll@{}}
        \toprule
        \textbf{No} & \textbf{Mask ID} & \textbf{Description (example)}              \\ 
        \midrule
        1   &  \texttt{ID\_IP\_ADDRESS}     &  IPv4 address (xxx.xxx.xxx.xxx)      \\
        2   &  \texttt{ID\_EMAIL}           &  Email address (xxx@yyy.zzz)       \\
        3   &  \texttt{ID\_ONION\_URL}      &  Onion URL                         \\
        4   &  \texttt{ID\_NORMAL\_URL}     &  Non-onion URL                     \\
        5   &  \texttt{ID\_BTC\_ADDRESS}    &  Bitcoin address                   \\
        6   &  \texttt{ID\_ETH\_ADDRESS}    &  Ethereum address                  \\
        7   &  \texttt{ID\_LTC\_ADDRESS}    &  Litecoin address                  \\
        8   &  \texttt{ID\_GENERAL\_MONEY}  &  Fiat money (10 USD, ¥500)         \\
        9   &  \texttt{ID\_CRYPTO\_MONEY}   &  Cryptocurrency (0.01 BTC, 10 mBTC) \\
        10  &  \texttt{ID\_WEIGHT}          &  Weight (10kg, 10lbs)              \\
        11  &  \texttt{ID\_LENGTH}          &  Length (10cm, 10mm)               \\
        12  &  \texttt{ID\_VOLUME}          &  Volume (10ml, 5L)                 \\
        13  &  \texttt{ID\_TIME}            &  \begin{tabular}[t]{@{}l@{}} 
                                         Widely used date/time format \\ 
                                         (2000-01-01 09:00:00, 01-Jan-2020)
                                      \end{tabular}                        \\
        14  &  \texttt{ID\_PERCENTAGE}      &  Percentage (50\%)                     \\
        15  &  \texttt{ID\_FILENAME}        &  \begin{tabular}[t]{@{}l@{}} 
                                         File names with popular extensions \\
                                         (xxx.zip, yyy.pdf)
                                      \end{tabular}                        \\
        16  &  \texttt{ID\_FILESIZE}        &  File size (10MB, 16GB)               \\
        17  &  \texttt{ID\_VERSION}         &  Version names (version 5.0, v1.0.0)  \\
        18  &  \texttt{ID\_NUMBER}          &  All the other number tokens          \\ 
        \bottomrule
    \end{tabular}%
    }
    \caption{Mask identifiers in {\dataset}}
    \label{tab:mask_id}
\end{table}

\subsection{Additional Processing \& Text Masking}
\label{sec:processing_masking}
As the Dark Web contains webpages in various languages, we label each of the documents in {\dataset} with the language of its content using fastText (\citealp{joulin2016fasttext}; \citealp{joulin2016bag}). To generalize unnecessary details and anonymize sensitive information in the Dark Web, we process each document by masking appropriate information with mask identifiers. A total of 18 types of mask identifiers are used to mask each document, as shown in Table \ref{tab:mask_id}. We utilize simple keywords, regular expressions, and a cryptocurrency address detection library\footnote{\url{https://pypi.org/project/cryptoaddress/}} to detect such phrases for masking. This prevents personal information such as email addresses from being included in our public dataset. Finally, to filter out noisy data and optimize text for linguistic analysis, we remove punctuations and words that are over 50 letters long, lemmatize words, and convert all text to lowercase.

\section{Data Analysis}
\label{sec:data_analysis}
%Before we classify the Dark Web documents into categories, we scrutinize the documents to boost our classification result. Because the Dark Web has unique characteristics (i.e., anonymity and volatility), we assumed they are also mirrored in the linguistic properties of the Dark Web. We believe that these linguistic properties are useful to figure out the topic of a document correctly. Then, we compare several metrics measuring the linguistic features extracted from the two Dark Web datasets ({\dataset} and DUTA) and the aggregate Surface Web dataset in order to report the hidden nature of the Dark Web language.

To assess the linguistic properties of the Dark Web, illustrate the characteristics of textual content in each category, and better understand the differences in the use of language in the Dark / Surface Web, we conduct an in-depth exploratory data analysis. We analyze the text data of {\dataset} and compare measurements with that of the other datasets (see Section \ref{sec:dataset_comparison}) as shown in Table \ref{tab:data-analysis}.\footnote{Some detailed methods and results obtained from our analyses are provided in Appendix \ref{sec:appendix-fig}.}

\begin{table}[t]
    \centering
    \resizebox{\columnwidth}{!}{%
        \begin{tabular}{@{}lccc@{}}
            \toprule
            \textbf{Data Analysis Methods / Statistics} & \multicolumn{3}{c}{\textbf{Dataset}}  \\ \cmidrule(l){2-4} 
                                               & \begin{tabular}[c]{@{}c@{}} {\dataset} \\ (ours)\end{tabular} & DUTA-10K & \begin{tabular}[c]{@{}c@{}}Surface Web \\ Aggregate\end{tabular}            \\ 
            \midrule
                                             \multicolumn{4}{c}{\textit{Analysis Using Raw Data}}                \\
            In-vocab / out-of-vocab words (\S \ref{sec:valid_word_ratio})          & $\circ$      & $\circ$  & $\circ$        \\
            PoS distribution (\S \ref{sec:pos-cf})                    & $\circ$      &          & $\circ$        \\
            Content / function word ratio (\S \ref{sec:pos-cf})       & $\circ$      &          & $\circ$        \\
            Word frequency distribution (\S \ref{sec:wordfreq_dist})  & $\circ$      &    $\circ$    & $\circ$        \\
            \midrule
                                             \multicolumn{4}{c}{\textit{Analysis Using Masked Data}}             \\ 
            Document similarity (\S \ref{sec:doc_sim})                & $\circ$      & $\circ$  &                \\
            Mask token distribution (\S \ref{sec:mask-tfidf})         & $\circ$      &          &                \\
            TF-IDF (\S \ref{sec:mask-tfidf})                          & $\circ$      &          &                \\
            \bottomrule
        \end{tabular}%
    }
    \caption{List of data analysis methods/statistics and comparisons between datasets. Some analyses are presented in the Appendix.}
    \label{tab:data-analysis}
\end{table}

\subsection{Datasets for Comparison}
\label{sec:dataset_comparison}
We compare {\dataset} with DUTA to look for any significant differences between the two Dark Web corpora, and use documents labeled as English for our analysis. We also aggregate three existing text datasets in English with Surface Web content (we consider each of these datasets as a single subcategory within the aggregate Surface Web data) from here on to compare between the Dark / Surface Web domains. These categories are chosen to encompass the various topics and language styles (formal / informal) used throughout the Surface Web.

The aggregate Surface Web dataset consists of the following categories: the IMDb Large Movie Review Dataset \citep{maas2011learning}, the Wikitext-2 Dataset \citep{merity2016pointer}, and the Reddit Corpus \citep{chang-etal-2020-convokit}, to represent review texts, wiki articles, and online forum discussions, respectively. To match the size of the dataset with its Dark Web counterparts, the aggregate Surface Web dataset is trimmed by randomly sampling a portion of documents from each category. The total raw word count of each dataset is shown in Table \ref{tab:wordstats}. 

It is worth noting that we use raw text data instead of the masked data for some analyses to reduce bias; for example, the Surface Web aggregate dataset is not masked, so we use the non-masked versions of the Dark Web datasets for some comparisons.

\begin{table}[t]
    \centering
    \resizebox{\columnwidth}{!}{%
    \begin{tabular}{@{}cccc@{}}
        \toprule
        \textbf{Dataset}      & \textbf{Total Word Count} & \textbf{In-vocab} & \textbf{Out-of-vocab} \\ \midrule
        {\dataset}            & 9.51M                     & \begin{tabular}[c]{@{}c@{}}39.6\%\\ (45670)\end{tabular} & \begin{tabular}[c]{@{}c@{}}60.4\%\\ (69696)\end{tabular} \\
        DUTA-10K              & 7.60M                     & \begin{tabular}[c]{@{}c@{}}44.7\%\\ (37676)\end{tabular} & \begin{tabular}[c]{@{}c@{}}55.3\%\\ (46651)\end{tabular} \\
        Surface Web Aggregate & 7.41M                     & \begin{tabular}[c]{@{}c@{}}44.9\%\\ (46817)\end{tabular} & \begin{tabular}[c]{@{}c@{}}55.1\%\\ (57361)\end{tabular} \\ \bottomrule
    \end{tabular}%
    }
    \caption{The total raw word count and the number of unique in-vocab / out-of-vocab words in each dataset. We only consider words that are sequences of purely alphabetic characters separated by whitespace.}
    \label{tab:wordstats}
\end{table}

\subsection{Analysis Methods \& Results}
\label{sec:analysis_methods}

\subsubsection{In-vocab / Out-of-vocab Word Analysis}
\label{sec:valid_word_ratio}

It is known that Dark Web users intentionally use obscure slangs and words to refer to specific items (\citealp{harviainen2020drug}; \citealp{zhang2020survey}). To verify if this behavior affects the types of words seen in the Dark Web, we analyze the \textit{in-vocabulary} and \textit{out-of-vocabulary} words by building a list of unique words in each dataset and determining the presence of each word in a spellchecker library\footnote{\url{https://github.com/barrust/pyspellchecker}}. Words not listed in the library's dictionary are defined to be \textit{out-of-vocabulary}. Note out-of-vocab words do not necessarily correspond to incorrect or nonexistent words; for example, there are many abbreviations that are out-of-vocab but are widely used online.

The results are shown in Table \ref{tab:wordstats}. Due to the limited number of in-vocab words in the dictionary, it follows that in-vocab word ratio decreases with higher total word counts. Therefore, not much can be said about the lexical characteristics of the Dark Web from the ratios.
To see which out-of-vocab words frequently appear in each corpus, we rank them by their frequency. In the Surface Web, common abbreviations, well-known companies, and celebrity names rank high in the list, while explicit slangs, malicious activity-specific abbreviations, and misspellings (cann\textit{a}balism, pedo\textit{f}ilia, s\textit{hc}ool) manifest the Dark Web. A sample comparison of abbreviations found in each web domain is shown in Table \ref{tab:common-abbr}. The Surface Web mainly exhibits commonly used abbreviations such as measurement units, household products, and colloquial Internet language, while the Dark Web exhibits abbreviations related to financial services, drugs, and pornography.

\begin{table}[t]
    \centering
    \resizebox{0.72\columnwidth}{!}{%
        \begin{tabular}{@{}ccc|ccc@{}}
            \toprule
            \multicolumn{6}{c}{\textbf{Common Abbreviations}}                           \\
            \multicolumn{3}{c}{\textbf{CoDA (Dark Web)}} & \multicolumn{3}{c}{\textbf{Surface Web}} \\ \midrule
            btc & cp  & cvc  & btw  & dvd  & idk  \\
            cvv & hd  & irc  & imo  & km   & lmao \\
            lsd & mg  & pthc & mph  & pc   & st   \\
            ssn & vpn & xxx  & tv   & vs   & wtf  \\ \bottomrule 
        \end{tabular}%
    }
    \caption{Some common abbreviation examples of {\dataset} and Surface Web Aggregate. These abbreviations do not show up on pyspellchecker's dictionary and are marked as out-of-vocab words.}
    \label{tab:common-abbr}
\end{table}

\subsubsection{Word Frequency Distribution}
\label{sec:wordfreq_dist}
It is well known that large text corpora tend to follow Zipf's law \citep{piantadosi2014zipf}, which states that the word frequency distribution is proportional to a power law of the form:
\begin{equation}
\nonumber
f(r) \propto r^{-\alpha}
\end{equation}
for $\alpha \approx 1$, where $r$ is the frequency rank of the word (most frequent word has $r=1$ and so on) and $f(r)$ is the frequency of a word with rank $r$. To verify whether the characteristics of language used in the Dark Web affect the power law distribution of words, we compare the word frequency distribution between Dark and Surface Web corpora. We aggregate all texts in each category into a single file, lemmatize each word using spaCy\footnote{\url{https://spacy.io/}}, and use scikit-learn\footnote{\url{https://scikit-learn.org/stable/}} \citep{scikit-learn} to retrieve the word frequency per category. 

We find that, as far as word frequency distribution is concerned, there is no significant difference between the Dark Web and the Surface Web\footnote{Word frequency distribution is shown in Figure \ref{fig:wordfreq_dist}.}. As the Dark Web contains many phishing sites which are near identical copies of each other \citep{yoon2019doppelgangers}, we believed that some words may have abnormally high frequencies, which would affect the overall distribution. However, the results suggest that word frequency distribution is largely domain-invariant. 

%We assumed that the Dark Web text shows somewhat different word frequency distribution from that of the Surface Web, but we see that word frequency distribution is domain-invariant.

\iffalse
\begin{figure}[t]
    \centering
    \includegraphics[width=\columnwidth,trim={0.2cm 0.2cm 0.2cm 0.5cm},clip]{figure/coda_dist_w_surf.pdf}
    \caption{Log-log plot of word frequency distribution of {\dataset} and Surface Web aggregate data by category}
    \label{fig:wordfreq_dist}
\end{figure}
\fi

\begin{figure}[t]
    \centering
    \includegraphics[width=\columnwidth,trim={0.2cm 0.2cm 0.2cm 1.3cm},clip]{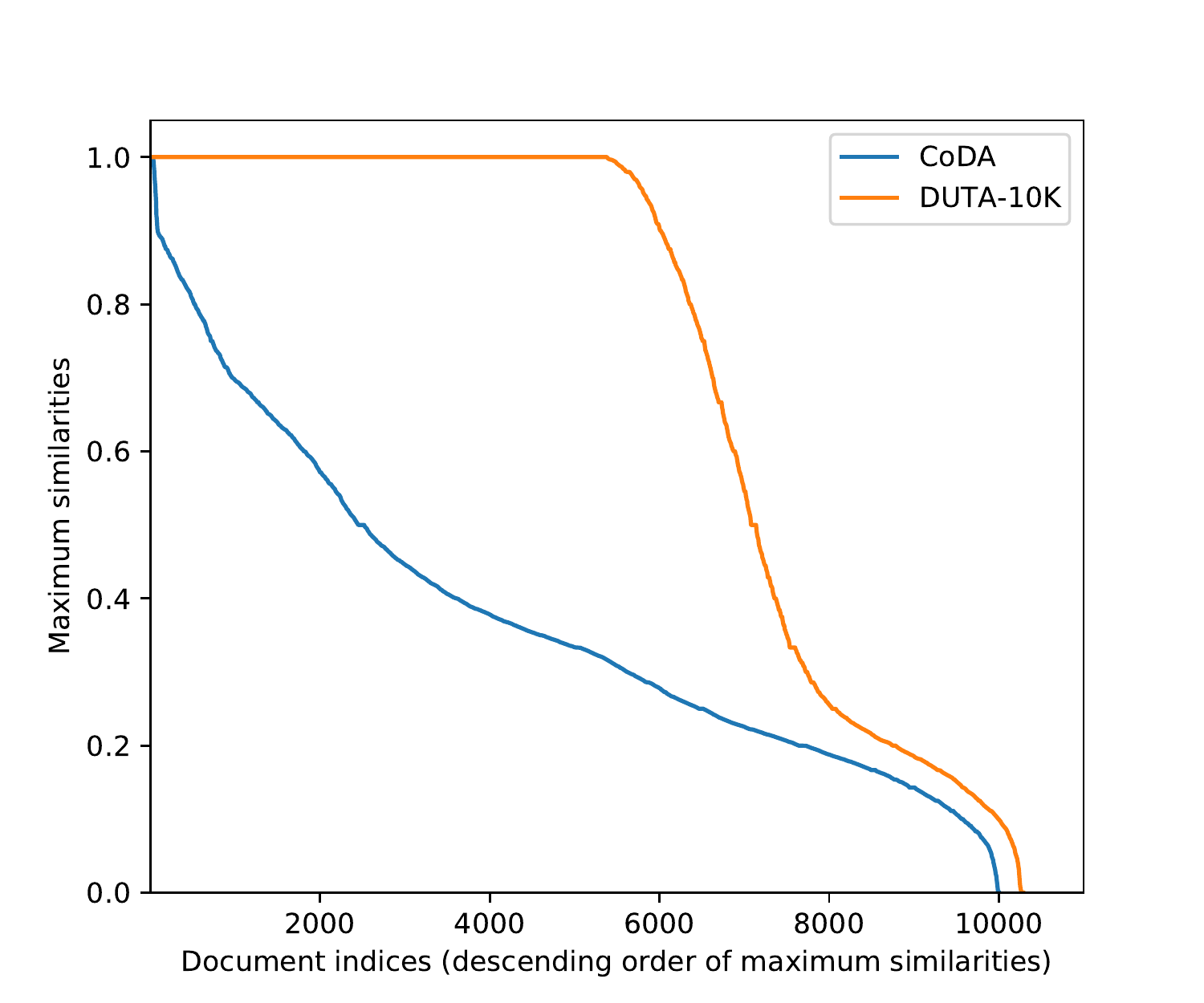}
    \caption{Maximum similarity (Jaccard distance) graph of documents in {\dataset} and DUTA. The similarities between every document in the same dataset are measured and the maximum similarity is taken for each document.}
    \label{fig:maxsim}
\end{figure}

\subsubsection{Document Similarity}
\label{sec:doc_sim}

As mentioned in Section \ref{sec:introduction}, about half of DUTA's documents contain duplicate data. {\dataset} addresses this problem by crawling web pages whose textual content is less similar to one another. To this end, we measure the document similarity between the two Dark Web corpora to validate the uniqueness of documents in {\dataset}. Through manual inspection, we find that some pages share the same exact content but with slight variations in details such as numbers. To prevent such differences from affecting the document similarity, we mask and preprocess documents in DUTA in the same manner as {\dataset} and convert each document into a bag of lowercase words. The similarity is measured by taking the Jaccard distance on the bags of words, with distance of 1 indicating complete similarity between two documents.

To illustrate the amount of overlapping content in {\dataset} and DUTA, we compare each document with all other documents from the same corpus, and denote the maximum Jaccard distance as its maximum similarity. As shown in Figure~\ref{fig:maxsim}, more than half of the documents in DUTA share almost completely overlapping content, whereas \dataset{} exhibits much lower similarities overall. Although DUTA consists of data from 10367 onion services which is larger than the number of onion services collected for use in \dataset{}, this shows that the data in \dataset{} is more uniquely varied and thus has higher information density.

\subsubsection{Mask ID Distribution and TF-IDF}
\label{sec:mask-tfidf}

To gain insight into some of the lexical characteristics of each category in the Dark Web, we evaluate the mask ID distribution and TF-IDF (term frequency-inverse document frequency) for {\dataset}. The mask ID distribution is calculated by dividing the frequency of a particular mask ID (listed in Table \ref{tab:mask_id}) in a document by the number of all mask IDs in that document (we exclude \texttt{ID\_NUMBER} in our data for this analysis as it accounts for the majority of all masks in every category). This is done for every document, and we take the average distribution by category. Similar methods are used for TF-IDF using scikit-learn \citep{scikit-learn}. We exclude English stopwords as defined in NLTK \citep{bird2009natural}, but preserve the mask IDs to capture important mask IDs in each category.

We find that some mask IDs are particularly representative in some categories. For example, the \textit{drugs} category has a high proportion of \texttt{ID\_WEIGHT}, which is reasonable as webpages in \textit{drugs} usually specify the weight of the drug in their listings. The TF-IDF measurements also show some interesting results\footnote{Detailed measurements are provided in Appendix \ref{sec:appendix-tfidf}.}; for example, the majority of terms with the highest TF-IDF in the \textit{electronics} category are related to Apple products (iPhones, iPads, MacBooks, etc.) which may suggest a high popularity of these products in the Dark Web.

\iffalse
\begin{figure}[t]
    \centering
    \includegraphics[width=\columnwidth,trim={0.2cm 0.2cm 0.2cm 0.5cm},clip]{figure/mask_dist_bar.pdf}
    \caption{Mask token distribution by category}
    \label{fig:mask_dist}
\end{figure}
\fi
\section{Classification Experiments}
\label{sec:classification_experiments}

\subsection{Setup}
We build several classifiers to investigate the performance of existing classification models on Dark Web text. Although deep neural network models are widely used today \citep{minaee2021deep}, simple machine learning models such as SVM and na\"ive Bayes (NB) have been reported to perform reasonably well on Dark Web texts, often outperforming deep models \citep{choshen-etal-2019-language}. Therefore, we evaluate both types of models to see which is adequate for Dark Web text classification. We split \dataset{} into training and test sets (7:3 ratio) after stratified random shuffling with the same random seed for all experiments. The preprocessing method used for document similarity (Section \ref{sec:doc_sim}) is applied here as it empirically works best across models.\footnote{Detailed training configurations of each classifier are provided in Appendix \ref{sec:experimental-details}.}

\noindent \textbf{Multi-class SVM:} We train a multi-class SVM classifier with TF-IDF features, and tune its hyperparameters by grid search. We build our classifier using \texttt{TfidfVectorizer}, \texttt{LinearSVC}, and \texttt{GridSearchCV} classes in scikit-learn.

\iffalse
\noindent \textbf{Bag of embeddings:} In addition to XXXXX models, we also built neural network-based classifiers. First, we started from a linear model that has a single fully connected layer and a CRITERION FUNCTION(CROSS ENTROPY LOSS). Its embedding and hidden size is XXXXX and YYYYY, respectively. Like multi-class SVM model, the linear model shows pretty good results without making a lot of effort.
\fi

\noindent \textbf{CNN:} Convolutional Neural Networks have been established as one of the popular choices for text classification for the ability to recognize position-invariant patterns such as text phrases \citep{minaee2021deep}. Using PyTorch, we build a CNN model with a GloVe embedding layer (6B.300d), 2D convolution layers, and a fully-connected layer \citep{pennington2014glove}.

\noindent \textbf{BERT:} To benefit from contextual representations and transfer learning, we use BERT \citep{devlin-etal-2019-bert}, a state-of-the-art language model widely adopted across many NLP and machine learning tasks. We use the pretrained \texttt{bert-base-uncased} model in the PyTorch version of the HuggingFace library \citep{wolf-etal-2020-transformers} with a fully-connected classification layer on top of the \texttt{[CLS]} token.

\subsection{Results}
Table~\ref{tab:result} summarizes the performance of the three classifiers on CoDA. BERT exhibits the best results possibly due to its capability to model unknown words and utilize contextual information, despite the lexical differences of the Dark Web as shown in Section \ref{sec:analysis_methods}. SVM produces comparable results, suggesting that the relatively simple bag-of-words approach is still very effective at modeling topics of such domain-specific text. In contrast, CNN fares relatively worse, which is likely due to the specialized vocabulary of the Dark Web being poorly covered by the pretrained word embedding as seen in \citet{choshen-etal-2019-language}.

Table~\ref{tab:result-detail} shows the detailed results of classification using BERT. The classifier works relatively well for categories that exhibit a smaller specialized vocabulary such as \textit{arms}, \textit{electronics}, and \textit{gambling}, whereas it performs worse for categories that cover diverse subtopics such as \textit{cryptocurrency} and \textit{financial}. We also observe that the classifier often confuses \textit{hacking} with \textit{cryptocurrency} or \textit{financial} especially when documents contain phrases such as ``hacked PayPal'' or ``hacked Bitcoin wallets'', which are not categorized in the \textit{hacking} category by our guidelines (the \textit{hacking} category relates to hacking services and professional hacking techniques).

\begin{table}[t]
    \centering
    \resizebox{0.7\columnwidth}{!}{%
    \begin{tabular}{@{}lccc@{}}
        \toprule
               & \textbf{Precision} & \textbf{Recall} & \textbf{F1-score} \\ 
        \midrule
        SVM    &  91.59  &  91.17  &  91.19  \\
        CNN    &  88.08  &  87.30  &  87.23  \\
        BERT   &  \textbf{92.51}  &  \textbf{92.50}  &  \textbf{92.49}  \\
        \bottomrule
    \end{tabular}%
    }
    \caption{Classifier performance on CoDA (weighted avg.). Boldface represents best performance.}
    \label{tab:result}
\end{table}

\begin{table}[t]
    \centering
    \resizebox{0.8\columnwidth}{!}{%
    \begin{tabular}{@{}lccc@{}}
        \toprule
        \textbf{Category} & \textbf{Precision} & \textbf{Recall} & \textbf{F1-score} \\ 
        \midrule
        Arms / Weapons &  96.70  &  97.78  &  97.24  \\
        Cryptocurrency &  90.45  &  87.28  &  88.84  \\
        Drugs          &  93.90  &  92.29  &  93.08  \\
        Electronics    &  94.66  &  96.88  &  95.75  \\
        Financial      &  90.71  &  94.02  &  92.33  \\
        Gambling       &  99.15  &  98.31  &  98.72  \\
        Hacking        &  87.50  &  89.74  &  88.61  \\
        Pornography    &  94.20  &  94.46  &  94.33  \\
        Violence       &  93.15  &  93.79  &  93.47  \\
        Others         &  90.45  &  89.73  &  90.09  \\
        \midrule
        \textbf{Weighted avg.} & \textbf{92.51} & \textbf{92.50} & \textbf{92.49} \\
        \bottomrule
    \end{tabular}%
    }
    \caption{Per-category performance of BERT on CoDA}
    \label{tab:result-detail}
\end{table}

\section{Use Cases}
\label{sec:usecase}
In this section, we elaborate on the use cases of our corpus and the classifiers trained on {\dataset} and DUTA.

\noindent \textbf{(1) Synonym Inference:} Dark Web users tend to use words differently from their original meaning to conceal or disguise their intents. For example, we observed that car company names (e.g., \textit{Tesla}, \textit{Toyota}) are often used in drug-related documents in the Dark Web to refer to synthetic drugs with brand logos imprinted on each pill.

We test the above scenario by training two simple Word2vec models~\cite{word2vec, rehurek2010software}, one using {\dataset} and another using DUTA. For each model, we query \textit{Tesla} and \textit{Toyota} and retrieve the most similar words to the queried terms. In this case, both models output drug-related words such as \textit{methoxphenidine}, \textit{testosterone}, and \textit{alprazolam}. We also query another word, \textit{Wasabi}, which originally refers to a plant but is also used to refer to a Bitcoin wallet service. In the Dark Web, \textit{Wasabi} is more likely to be used as a cryptocurrency term rather than the plant itself. When \textit{Wasabi} is queried, the model trained on {\dataset} returns cryptocurrency-related words, while the model trained on DUTA does not have the word in its vocabulary. We list the top 10 most similar words to \textit{Wasabi} as reported by the model trained on {\dataset}, most of which are related to cryptocurrency services: \textit{mustard, electrum, samourai, pools, trustless, mycelium, rpc, xapo, hijacker, converter}.

%WASABI, ETHEREUM, TESLA, ICE, MOUSE, GLOCK, KUSH

\noindent \textbf{(2) Topic Classification:} We assess the efficacy of {\dataset} and DUTA in classifying document categories in the Dark Web. For this scenario, we manually compile a list of 34 forum / marketplace websites on the Dark Web across three different topics: \textit{drugs}, \textit{weapons}, and \textit{finance}, and create an extra benchmark dataset consisting of 2236 webpages from the list\footnote{Refer to Appendix \ref{sec:appendix-usecase2} for the full list of website names and URLs.}. To remove possible overlap between the {\dataset} / DUTA corpora and the benchmark dataset, we exclude any documents crawled from the same URL as those from the benchmark dataset, or documents that mention any of the names from the benchmark websites in their content. This excludes 246 and 220 documents from {\dataset} and DUTA, respectively. We then train a BERT-based classifier for each Dark Web corpus on the remaining documents with the same configuration used in the classification experiments, and evaluate them on the benchmark dataset.\footnote{Since DUTA uses a different category set, we employ the following category mapping: \textit{counterfeit-credit-cards}, \textit{counterfeit-money}, and \textit{counterfeit-personal-identification} to finance, and \textit{drugs} and \textit{violence} to drugs and weapons categories, respectively.}

\begin{table}[t]
    \centering
    \resizebox{\columnwidth}{!}{%
    \begin{tabular}{@{}llcccc@{}}
        \toprule
        \textbf{Model} & \textbf{Category} & \textbf{\# pages} & \textbf{Precision} & \textbf{Recall} & \textbf{F1-score} \\ 
        \midrule
        {\dataset}  &  Drugs    &  936   &  99.87   &  83.44  &  90.92 \\
                    &  Weapons  &  674   &  100.00  &  98.37  &  99.18 \\
                    &  Finance  &  626   &  96.96   &  76.36  &  85.43 \\
                    &  \textbf{All} & \textbf{2236} & \textbf{99.09} & \textbf{85.96} & \textbf{91.87} \\
        \midrule
        DUTA  &  Drugs    &  936   &  99.46   &  78.74  &  87.90 \\
              &  Weapons  &  674   &  99.84   &  94.96  &  97.34 \\
              &  Finance  &  626   &  98.89   &  71.09  &  82.71 \\
              &  \textbf{All} & \textbf{2236} & \textbf{99.42} & \textbf{81.48} & \textbf{89.29} \\ 
        \bottomrule
    \end{tabular}%
    }
    \caption{Classification performance of the BERT-based classifier on the benchmark dataset}
    \label{tab:benchmark_perf}
\end{table}

Table \ref{tab:benchmark_perf} shows the classification performance measured on the benchmark dataset, in which the {\dataset}-trained classifier consistently outperforms the DUTA-trained classifier. We conjecture that this is because {\dataset} contains less duplicate text with more diverse domain-specific words in the same number of documents, allowing the trained classifier to generalize better to unseen documents.

\section{Conclusion}
\label{sec:conclusion}
In this work, we introduced {\dataset}, a Dark Web text corpus collected from various onion services divided into topical categories. Using {\dataset}, we conducted a thorough analysis of the linguistic properties of the Dark Web and found that there are clear lexical differences from the Surface Web including abbreviations and lexical structure such as PoS distribution. We also found lexical characteristics of categories through mask ID distribution and TF-IDF.

Our text classification results showed that SVM and BERT perform well in the Dark Web domain, even with the language differences that the Dark Web exhibits compared to that of the Surface Web. Finally, we have demonstrated the practicality of {\dataset} through two use cases with NLP methods. We speculate that the lack of duplicate content in {\dataset} compared to DUTA may aid in the performance of such applications. 

We hope that our dataset and our work motivates further research in the field of language-based Dark Web analysis. 

\section*{Ethical Considerations}
\subsection*{Masking Sensitive Information}
Due to the nature of anonymous networks such as Tor, raw data posted on the Dark Web may contain private or illegal information. Such information includes (but is not limited to) Bitcoin addresses, credit card information, and social security numbers. Since {\dataset} was compiled by randomly selecting web documents from the Dark Web, the dataset may contain such information. Therefore, it is important that a public Dark Web dataset such as {\dataset} addresses ethical issues regarding sensitive information.

To prevent the use of {\dataset} for malicious purposes such as the extraction of sensitive information, we identify types of potentially sensitive data (such as email, IP, URL, crypto addresses, and social security numbers) which are subsequently masked (refer to Section \ref{sec:processing_masking} for the detailed description on mask identifiers used). These identifiers are matched using regular expressions and each page has been manually double-checked on whether such content has been properly masked by the authors. During this time, the authors did not find sensitive content outside of the masked information.

As mentioned in Section \ref{sec:data_analysis}, some data analysis methods are conducted with unmasked versions of the Dark Web to prevent bias. However, we only use the unmasked version of {\dataset} for a fair analysis between the Dark and the Surface Web data, and do not utilize or disclose information found in the unmasked data in any way.

\subsection*{Handling Illegal Content}

A significant portion of the Dark Web deals with explicit, pornographic content (violence, child pornography, torture, etc.). The act of accessing or viewing such media is illegal by law in many parts of the world. To prevent the access of such media, we collect crawled Dark Web pages in the form of HTML and parse HTML tags to retrieve only the text data. In addition, URL addresses and onion addresses that may link to such illegal media are also masked as previously mentioned (note that all URL addresses and onion addresses have been masked, regardless of their content). Consequently, the authors and the annotators do not have access to media that are illegal by law.

It is worth noting that {\dataset} still contains texts of various activities that occur in the Dark Web, some of which are illegal in nature (drug trade, counterfeit products, etc.). However, the inclusion of text in the dataset that describes potentially illegal activities is not of ethical concern. Therefore, we do not censor text data that correspond to illegal activities.

\subsection*{Ethics on Annotation}

Dark Web content often contains sensitive and illicit activities. Dealing with such content during the annotation process may be unsettling for some people, so we chose annotators who are experienced with the Dark Web and has given consent to being exposed to such content. The annotators recruited for classifying {\dataset} were specialists who work at a cyber threat data analytics \& intelligence company specializing in Dark Web data. To ensure that the annotation process is fair, each of the ten annotators handled the same number of pages and were given equal compensations.

\subsection*{Preventative Measures to Discourage Non-Academic Use of CoDA}

The content of CoDA includes text descriptions of various Dark Web activities. A potential harm of releasing this dataset is bringing increased attention to these activities. We strongly believe that our research should be used for scientific purposes only and discourage non-academic use of CoDA. We take a preventative approach by only permitting access to CoDA to researchers with research purposes that abide by the ACL Code of Ethics. The terms of use agreement can be found at \url{https://s2w.inc/resources/coda}.
\section*{Acknowledgements}
This research was supported by the Engineering Research Center Program through the National Research Foundation of Korea (NRF) funded by the Korean Government MSIT (NRF-2018R1A5A1059921).

% \section*{Acknowledgments}

% The acknowledgments should go immediately before the references. Do not number the acknowledgments section.
% \textbf{Do not include this section when submitting your paper for review.}

\balance
\bibliographystyle{acl_natbib}
\bibliography{anthology,custom}

\begin{thebibliography}{36}
\expandafter\ifx\csname natexlab\endcsname\relax\def\natexlab#1{#1}\fi

\bibitem[{Al-Nabki et~al.(2019)Al-Nabki, Fidalgo, Alegre, and
  Fernández-Robles}]{al2019torank}
Mhd~Wesam Al-Nabki, Eduardo Fidalgo, Enrique Alegre, and Laura
  Fernández-Robles. 2019.
\newblock \href {https://doi.org/https://doi.org/10.1016/j.eswa.2019.01.029}
  {Torank: Identifying the most influential suspicious domains in the tor
  network}.
\newblock \emph{Expert Systems with Applications}, 123:212--226.

\bibitem[{Al~Nabki et~al.(2017)Al~Nabki, Fidalgo, Alegre, and
  de~Paz}]{al2017classifying}
Mhd~Wesam Al~Nabki, Eduardo Fidalgo, Enrique Alegre, and Ivan de~Paz. 2017.
\newblock \href {https://aclanthology.org/E17-1004} {Classifying illegal
  activities on tor network based on web textual contents}.
\newblock In \emph{Proceedings of the 15th Conference of the {E}uropean Chapter
  of the Association for Computational Linguistics: Volume 1, Long Papers},
  pages 35--43, Valencia, Spain. Association for Computational Linguistics.

\bibitem[{Artstein(2017)}]{artstein2017kappa}
Ron Artstein. 2017.
\newblock \emph{Handbook of linguistic annotation}.
\newblock Springer, Dordrecht.

\bibitem[{Avarikioti et~al.(2018)Avarikioti, Brunner, Kiayias, Wattenhofer, and
  Zindros}]{avarikioti2018structure}
Georgia Avarikioti, Roman Brunner, Aggelos Kiayias, Roger Wattenhofer, and
  Dionysis Zindros. 2018.
\newblock \href {https://doi.org/10.48550/ARXIV.1811.01348} {Structure and
  content of the visible darknet}.

\bibitem[{Bird et~al.(2009)Bird, Klein, and Loper}]{bird2009natural}
Steven Bird, Ewan Klein, and Edward Loper. 2009.
\newblock \emph{Natural language processing with Python: analyzing text with
  the natural language toolkit}.
\newblock O'Reilly Media, Inc.

\bibitem[{Biryukov et~al.(2014)Biryukov, Pustogarov, Thill, and
  Weinmann}]{biryukov2014content}
Alex Biryukov, Ivan Pustogarov, Fabrice Thill, and Ralf-Philipp Weinmann. 2014.
\newblock \href {https://doi.org/10.1109/ICDCSW.2014.20} {Content and
  popularity analysis of tor hidden services}.
\newblock In \emph{2014 IEEE 34th International Conference on Distributed
  Computing Systems Workshops (ICDCSW)}, pages 188--193.

\bibitem[{Chang et~al.(2020)Chang, Chiam, Fu, Wang, Zhang, and
  Danescu-Niculescu-Mizil}]{chang-etal-2020-convokit}
Jonathan~P. Chang, Caleb Chiam, Liye Fu, Andrew Wang, Justine Zhang, and
  Cristian Danescu-Niculescu-Mizil. 2020.
\newblock \href {https://www.aclweb.org/anthology/2020.sigdial-1.8}
  {{C}onvo{K}it: A toolkit for the analysis of conversations}.
\newblock In \emph{Proceedings of the 21th Annual Meeting of the Special
  Interest Group on Discourse and Dialogue}, pages 57--60, 1st virtual meeting.
  Association for Computational Linguistics.

\bibitem[{Chen(2011)}]{chen2011dark}
Hsinchun Chen. 2011.
\newblock \emph{Dark web: Exploring and data mining the dark side of the web},
  volume~30.
\newblock Springer Science \& Business Media.

\bibitem[{Choshen et~al.(2019)Choshen, Eldad, Hershcovich, Sulem, and
  Abend}]{choshen-etal-2019-language}
Leshem Choshen, Dan Eldad, Daniel Hershcovich, Elior Sulem, and Omri Abend.
  2019.
\newblock \href {https://doi.org/10.18653/v1/P19-1419} {The language of legal
  and illegal activity on the {D}arknet}.
\newblock In \emph{Proceedings of the 57th Annual Meeting of the Association
  for Computational Linguistics}, pages 4271--4279, Florence, Italy.
  Association for Computational Linguistics.

\bibitem[{Devlin et~al.(2019)Devlin, Chang, Lee, and
  Toutanova}]{devlin-etal-2019-bert}
Jacob Devlin, Ming-Wei Chang, Kenton Lee, and Kristina Toutanova. 2019.
\newblock \href {https://doi.org/10.18653/v1/N19-1423} {{BERT}: Pre-training of
  deep bidirectional transformers for language understanding}.
\newblock In \emph{Proceedings of the 2019 Conference of the North {A}merican
  Chapter of the Association for Computational Linguistics: Human Language
  Technologies, Volume 1 (Long and Short Papers)}, pages 4171--4186,
  Minneapolis, Minnesota. Association for Computational Linguistics.

\bibitem[{Ghosh et~al.(2017)Ghosh, Das, Porras, Yegneswaran, and
  Gehani}]{ghosh2017automated}
Shalini Ghosh, Ariyam Das, Phil Porras, Vinod Yegneswaran, and Ashish Gehani.
  2017.
\newblock \href {https://doi.org/10.1145/3097983.3098193} {Automated
  categorization of onion sites for analyzing the darkweb ecosystem}.
\newblock KDD '17, page 1793–1802, New York, NY, USA. Association for
  Computing Machinery.

\bibitem[{Guitton(2013)}]{guitton2013review}
Clement Guitton. 2013.
\newblock \href {https://doi.org/https://doi.org/10.1016/j.chb.2013.07.031} {A
  review of the available content on tor hidden services: The case against
  further development}.
\newblock \emph{Computers in Human Behavior}, 29(6):2805--2815.

\bibitem[{Harviainen et~al.(2020)Harviainen, Haasio, and
  H\"{a}m\"{a}l\"{a}inen}]{harviainen2020drug}
J.~Tuomas Harviainen, Ari Haasio, and Lasse H\"{a}m\"{a}l\"{a}inen. 2020.
\newblock \href {https://doi.org/10.1145/3377290.3377293} {Drug traders on a
  local dark web marketplace}.
\newblock AcademicMindtrek '20, page 20–26, New York, NY, USA. Association
  for Computing Machinery.

\bibitem[{He et~al.(2019)He, He, and Li}]{he2019classification}
Siyu He, Yongzhong He, and Mingzhe Li. 2019.
\newblock \href {https://doi.org/10.1145/3322645.3322691} {Classification of
  illegal activities on the dark web}.
\newblock In \emph{Proceedings of the 2019 2nd International Conference on
  Information Science and Systems}, ICISS 2019, page 73–78, New York, NY,
  USA. Association for Computing Machinery.

\bibitem[{Joulin et~al.(2016{\natexlab{a}})Joulin, Grave, Bojanowski, Douze,
  J{\'{e}}gou, and Mikolov}]{joulin2016fasttext}
Armand Joulin, Edouard Grave, Piotr Bojanowski, Matthijs Douze, Herv{\'{e}}
  J{\'{e}}gou, and Tom{\'{a}}s Mikolov. 2016{\natexlab{a}}.
\newblock \href {http://arxiv.org/abs/1612.03651} {Fasttext.zip: Compressing
  text classification models}.
\newblock \emph{CoRR}, abs/1612.03651.

\bibitem[{Joulin et~al.(2016{\natexlab{b}})Joulin, Grave, Bojanowski, and
  Mikolov}]{joulin2016bag}
Armand Joulin, Edouard Grave, Piotr Bojanowski, and Tom{\'{a}}s Mikolov.
  2016{\natexlab{b}}.
\newblock \href {http://arxiv.org/abs/1607.01759} {Bag of tricks for efficient
  text classification}.
\newblock \emph{CoRR}, abs/1607.01759.

\bibitem[{Landis and Koch(1977)}]{avarikioti1977kappa}
J.~Richard Landis and Gary~G. Koch. 1977.
\newblock \href {http://www.jstor.org/stable/2529310} {The measurement of
  observer agreement for categorical data}.
\newblock \emph{Biometrics}, 33(1):159--174.

\bibitem[{Lee et~al.(2019)Lee, Yoon, Kang, Kim, Kim, Han, Son, and
  Shin}]{lee2019cybercriminal}
Seunghyeon Lee, Changhoon Yoon, Heedo Kang, Yeonkeun Kim, Yongdae Kim, Dongsu
  Han, Sooel Son, and Seungwon Shin. 2019.
\newblock \href {https://doi.org/10.14722/ndss.2019.23055} {Cybercriminal
  minds: An investigative study of cryptocurrency abuses in the dark web}.
\newblock In \emph{26th Annual Network and Distributed System Security
  Symposium (NDSS 2019)}.

\bibitem[{Maas et~al.(2011)Maas, Daly, Pham, Huang, Ng, and
  Potts}]{maas2011learning}
Andrew~L. Maas, Raymond~E. Daly, Peter~T. Pham, Dan Huang, Andrew~Y. Ng, and
  Christopher Potts. 2011.
\newblock \href {https://aclanthology.org/P11-1015} {Learning word vectors for
  sentiment analysis}.
\newblock In \emph{Proceedings of the 49th Annual Meeting of the Association
  for Computational Linguistics: Human Language Technologies}, pages 142--150,
  Portland, Oregon, USA. Association for Computational Linguistics.

\bibitem[{McCoy et~al.(2008)McCoy, Bauer, Grunwald, Kohno, and
  Sicker}]{mccoy2008shining}
Damon McCoy, Kevin Bauer, Dirk Grunwald, Tadayoshi Kohno, and Douglas Sicker.
  2008.
\newblock Shining light in dark places: Understanding the tor network.
\newblock In \emph{Privacy Enhancing Technologies}, pages 63--76, Berlin,
  Heidelberg. Springer Berlin Heidelberg.

\bibitem[{Meland et~al.(2020)Meland, Bayoumy, and Sindre}]{MELAND2020101762}
Per~Håkon Meland, Yara Fareed~Fahmy Bayoumy, and Guttorm Sindre. 2020.
\newblock \href {https://doi.org/https://doi.org/10.1016/j.cose.2020.101762}
  {The ransomware-as-a-service economy within the darknet}.
\newblock \emph{Computers \& Security}, 92:101762.

\bibitem[{Merity et~al.(2016)Merity, Xiong, Bradbury, and
  Socher}]{merity2016pointer}
Stephen Merity, Caiming Xiong, James Bradbury, and Richard Socher. 2016.
\newblock \href {https://doi.org/10.48550/ARXIV.1609.07843} {Pointer sentinel
  mixture models}.

\bibitem[{Mikolov et~al.(2013)Mikolov, Sutskever, Chen, Corrado, and
  Dean}]{word2vec}
Tomas Mikolov, Ilya Sutskever, Kai Chen, Greg~S Corrado, and Jeff Dean. 2013.
\newblock \href
  {https://proceedings.neurips.cc/paper/2013/file/9aa42b31882ec039965f3c4923ce901b-Paper.pdf}
  {Distributed representations of words and phrases and their
  compositionality}.
\newblock In \emph{Advances in Neural Information Processing Systems},
  volume~26. Curran Associates, Inc.

\bibitem[{Minaee et~al.(2021)Minaee, Kalchbrenner, Cambria, Nikzad, Chenaghlu,
  and Gao}]{minaee2021deep}
Shervin Minaee, Nal Kalchbrenner, Erik Cambria, Narjes Nikzad, Meysam
  Chenaghlu, and Jianfeng Gao. 2021.
\newblock \href {https://doi.org/10.1145/3439726} {Deep learning--based text
  classification: A comprehensive review}.
\newblock \emph{ACM Comput. Surv.}, 54(3).

\bibitem[{Moore and Rid(2016)}]{moore2016cryptopolitik}
Daniel Moore and Thomas Rid. 2016.
\newblock \href {https://doi.org/10.1080/00396338.2016.1142085} {Cryptopolitik
  and the darknet}.
\newblock \emph{Survival}, 58(1):7--38.

\bibitem[{Pedregosa et~al.(2011)Pedregosa, Varoquaux, Gramfort, Michel,
  Thirion, Grisel, Blondel, Prettenhofer, Weiss, Dubourg, Vanderplas, Passos,
  Cournapeau, Brucher, Perrot, and Duchesnay}]{scikit-learn}
F.~Pedregosa, G.~Varoquaux, A.~Gramfort, V.~Michel, B.~Thirion, O.~Grisel,
  M.~Blondel, P.~Prettenhofer, R.~Weiss, V.~Dubourg, J.~Vanderplas, A.~Passos,
  D.~Cournapeau, M.~Brucher, M.~Perrot, and E.~Duchesnay. 2011.
\newblock Scikit-learn: Machine learning in {P}ython.
\newblock \emph{Journal of Machine Learning Research}, 12:2825--2830.

\bibitem[{Pennington et~al.(2014)Pennington, Socher, and
  Manning}]{pennington2014glove}
Jeffrey Pennington, Richard Socher, and Christopher~D Manning. 2014.
\newblock {GloVe}: Global vectors for word representation.
\newblock In \emph{Proceedings of the 2014 conference on empirical methods in
  natural language processing (EMNLP)}, pages 1532--1543.

\bibitem[{Peters et~al.(2018)Peters, Neumann, Iyyer, Gardner, Clark, Lee, and
  Zettlemoyer}]{peters2018deep}
Matthew~E Peters, Mark Neumann, Mohit Iyyer, Matt Gardner, Christopher Clark,
  Kenton Lee, and Luke Zettlemoyer. 2018.
\newblock Deep contextualized word representations.
\newblock \emph{arXiv preprint arXiv:1802.05365}.

\bibitem[{Piantadosi(2014)}]{piantadosi2014zipf}
Steven~T Piantadosi. 2014.
\newblock Zipf’s word frequency law in natural language: A critical review
  and future directions.
\newblock \emph{Psychonomic bulletin \& review}, 21(5):1112--1130.

\bibitem[{Ranaldi et~al.(2022)Ranaldi, Nourbakhsh, Patrizi, Ruzzetti, Onorati,
  Fallucchi, and Zanzotto}]{darksidelang2022}
Leonardo Ranaldi, Aria Nourbakhsh, Arianna Patrizi, Elena~Sofia Ruzzetti, Dario
  Onorati, Francesca Fallucchi, and Fabio~Massimo Zanzotto. 2022.
\newblock \href {https://doi.org/10.48550/ARXIV.2201.05613} {The dark side of
  the language: Pre-trained transformers in the darknet}.

\bibitem[{Rehurek and Sojka(2010)}]{rehurek2010software}
Radim Rehurek and Petr Sojka. 2010.
\newblock Software framework for topic modelling with large corpora.
\newblock In \emph{Proceedings of the LREC 2010 Workshop on New Challenges for
  NLP Frameworks}, pages 45--50.

\bibitem[{Sanchez-Rola et~al.(2017)Sanchez-Rola, Balzarotti, and
  Santos}]{sanchez2017onions}
Iskander Sanchez-Rola, Davide Balzarotti, and Igor Santos. 2017.
\newblock \href {https://doi.org/10.1145/3038912.3052657} {The onions have
  eyes: A comprehensive structure and privacy analysis of tor hidden services}.
\newblock In \emph{Proceedings of the 26th International Conference on World
  Wide Web}, WWW '17, page 1251–1260, Republic and Canton of Geneva, CHE.
  International World Wide Web Conferences Steering Committee.

\bibitem[{Soska and Christin(2015)}]{190886}
Kyle Soska and Nicolas Christin. 2015.
\newblock \href
  {https://www.usenix.org/conference/usenixsecurity15/technical-sessions/presentation/soska}
  {Measuring the longitudinal evolution of the online anonymous marketplace
  ecosystem}.
\newblock In \emph{24th {USENIX} Security Symposium ({USENIX} Security 15)},
  pages 33--48, Washington, D.C. {USENIX} Association.

\bibitem[{Wolf et~al.(2020)Wolf, Debut, Sanh, Chaumond, Delangue, Moi, Cistac,
  Rault, Louf, Funtowicz, Davison, Shleifer, von Platen, Ma, Jernite, Plu, Xu,
  Le~Scao, Gugger, Drame, Lhoest, and Rush}]{wolf-etal-2020-transformers}
Thomas Wolf, Lysandre Debut, Victor Sanh, Julien Chaumond, Clement Delangue,
  Anthony Moi, Pierric Cistac, Tim Rault, Remi Louf, Morgan Funtowicz, Joe
  Davison, Sam Shleifer, Patrick von Platen, Clara Ma, Yacine Jernite, Julien
  Plu, Canwen Xu, Teven Le~Scao, Sylvain Gugger, Mariama Drame, Quentin Lhoest,
  and Alexander Rush. 2020.
\newblock \href {https://doi.org/10.18653/v1/2020.emnlp-demos.6} {Transformers:
  State-of-the-art natural language processing}.
\newblock In \emph{Proceedings of the 2020 Conference on Empirical Methods in
  Natural Language Processing: System Demonstrations}, pages 38--45, Online.
  Association for Computational Linguistics.

\bibitem[{Yoon et~al.(2019)Yoon, Kim, Kim, Shin, and
  Son}]{yoon2019doppelgangers}
Changhoon Yoon, Kwanwoo Kim, Yongdae Kim, Seungwon Shin, and Sooel Son. 2019.
\newblock \href {https://doi.org/10.1145/3308558.3313551} {Doppelg\"{a}ngers on
  the dark web: A large-scale assessment on phishing hidden web services}.
\newblock In \emph{The World Wide Web Conference}, WWW '19, page 2225–2235,
  New York, NY, USA. Association for Computing Machinery.

\bibitem[{Zhang and Zou(2020)}]{zhang2020survey}
Hengrui Zhang and Futai Zou. 2020.
\newblock \href {https://doi.org/10.1109/ICCC51575.2020.9345271} {A survey of
  the dark web and dark market research}.
\newblock In \emph{2020 IEEE 6th International Conference on Computer and
  Communications (ICCC)}, pages 1694--1705.

\end{thebibliography}

\onecolumn

\newpage
\appendix
\label{sec:appendix}

\section{Inter-Annotator Agreement}
\label{sec:appendix-annotator}

To obtain high-quality annotations, we ran a training session with the first 150 documents in three 50-document intervals. Each stage, we asked 10 annotators to annotate the same documents with the guidelines, and measured inter-annotator agreements using Fleiss' Kappa coefficient~(\citealp{avarikioti1977kappa}; \citealp{artstein2017kappa}). After each interval, we held a discussion session to resolve disagreements and revised the guidelines to accommodate feedback from the annotators. For each interval, the agreement coefficients were 0.67, 0.72, and 0.88. Note that the coefficients greater than 0.60 and 0.80 can be interpreted as ``substantial'' and ``almost perfect'' agreements, respectively \citep{avarikioti1977kappa}. This suggests that the training sessions helped the annotators gradually reach a common consensus and familiarize themselves with the guidelines. We then assigned the remaining documents so that each document is assigned to a single annotator to speed up the annotation. The whole process took about three months, including one month of the training session. 

\section{Experimental Details}
\label{sec:experimental-details}

The data analysis experiments were performed on a machine with Intel Xeon E5-2630 v4 CPU @ 2.2 GHz (with no GPU usage), and the classification experiments were performed on a machine with Intel Xeon Gold 6258R CPU @ 2.70 GHz and Nvidia GeForce RTX 3090.

\textbf{SVM:} To fine-tune the parameters, we exhaustively generated all the candidate combinations of the two parameter pairs: tolerance and regularization. From a grid of their values, we applied 10-fold cross validation and found that the model with tolerance of 0.1 and regularization of 1.0 work best when we used `\texttt{balanced\_accuracy}' as the scoring strategy. Since this model is a multi-class classifier, we used the OVR (one-versus-rest) multi-class strategy. 

\textbf{CNN:} The model consists of one GloVe embedding layer (6B.300d), three 2-dimensional convolutions (Conv2d), and one fully-connected layer. The kernel sizes of the three convolution layers are 3, 4, and 5, respectively. We applied the ReLU activation function and 1-dimensional max pooling after each convolution layer. We also used the SGD optimizer and ran 10 training epochs with cross-entropy loss; the learning rate was 1.5, and the batch size was set to 32.

\textbf{BERT:} We used the Adam optimizer and ran 10 training epochs with cross-entropy loss, a learning rate of 2e-5, a linear schedule with no warmup step, a batch size of 32, and gradient norm clipping of 1.0. We also limited the maximum sequence length to 256 tokens, assuming that the leading part of text is indicative of topics. All the other settings are the same as those used in the original BERT paper.

\section{Language Distribution}
\label{sec:appendix-langdist}

We present the language distribution of {\dataset} in the following table:

\begin{table}[!ht]
    \centering
    \resizebox{0.75\columnwidth}{!}{%
    \begin{tabular}{@{}ll|ll|ll@{}}
        \toprule
        \textbf{Language} & \textbf{Document count} & \textbf{Language} & \textbf{Document count} & \textbf{Language} & \textbf{Document count} \\ 
        \midrule
        English    & 8855 & Persian    & 7 & Basque          & 1 \\
        Russian    & 542  & Swedish    & 7 & Egyptian Arabic & 1\\
        German     & 129  & Ukrainian  & 7 & Georgian        & 1\\
        French     & 100  & Turkish    & 6 & Greek           & 1\\
        Spanish    & 61   & Catalan    & 5 & Gujarati        & 1\\
        Portuguese & 54   & Hungarian  & 5 & Hindi           & 1\\
        Chinese    & 38   & Cebuano    & 3 & Ido             & 1\\
        Italian    & 28   & Esperanto  & 3 & Iloko           & 1\\
        Japanese   & 27   & Indonesian & 3 & Kurdish         & 1\\
        Dutch      & 14   & Latin      & 3 & Marathi         & 1\\
        Finnish    & 14   & Lithuanian & 3 & Punjabi         & 1\\
        Korean     & 12   & Norwegian  & 3 & Slovak          & 1\\
        Czech      & 11   & Bengali    & 2 & Tamil           & 1\\
        Polish     & 10   & Galician   & 2 & Thai            & 1\\
        Bulgarian  & 9    & Romanian   & 2 & Urdu            & 1\\
        Arabic     & 7    & Serbian    & 2 & Vietnamese      & 1\\
        Hebrew     & 7    & Slovenian  & 2 & Welsh           & 1\\
        \bottomrule
    \end{tabular}%
    }
    \caption{Language distribution of documents in {\dataset}}
    \label{tab:lang_dist}
\end{table}

\newpage

\section{Additional Data Analysis Methods \& Results}
\label{sec:appendix-fig}

Some additional details and figures of results collected from various data analyses methods in Section \ref{sec:data_analysis} are presented here.

\subsection{PoS Distribution \& Content Word / Function Word Ratio}
\label{sec:pos-cf}

\citet{choshen-etal-2019-language} demonstrated that legal and illegal texts in the Dark Web can be distinguishable through their lexical structure, that is, through part-of-speech (PoS) tags and distribution of content and function words. We analyze the PoS distribution and the distribution of content and function words in each dataset to see if there is a meaningful difference in the lexical structures of Dark / Surface Web contents. To obtain the universal PoS of words in the dataset, we utilize the PoS tagger in spaCy. Following \citet{choshen-etal-2019-language}, we define content words as words tagged by spaCy into to one of the following PoS tags:
\begin{equation}
\nonumber
    \{ \textsc{adj, adv, noun, propn, verb, x, num} \}
\end{equation}

\noindent and define all other words as function words. Since text length varies widely for each document in the Dark Web, we analyze the mean PoS ratio and the mean content word / function word ratio (CF ratio) for each category for both Dark and Surface Webs. The mean CF ratio $\bar{r}_{cf}(\mathcal{C})$ for some category $\mathcal{C}$ is given by 
\begin{equation}
\nonumber
    \bar{r}_{cf}(\mathcal{C}) = \frac{1}{|\mathcal{C}|}\sum_{d \in \mathcal{C}}\frac{N_{c}(d)}{N_{f}(d)}
\end{equation}

\noindent where $|\mathcal{C}|$ denotes the number of documents (text files) in category $\mathcal{C}$, and $N_{c}(d)$, $N_{f}(d)$ denote the total number of content words and function words in some document $d$, respectively. 

\begin{figure*}[!ht]
    \centering
    \includegraphics[width=\textwidth,trim={0.2cm 0.2cm 0.2cm 0.2cm},clip]{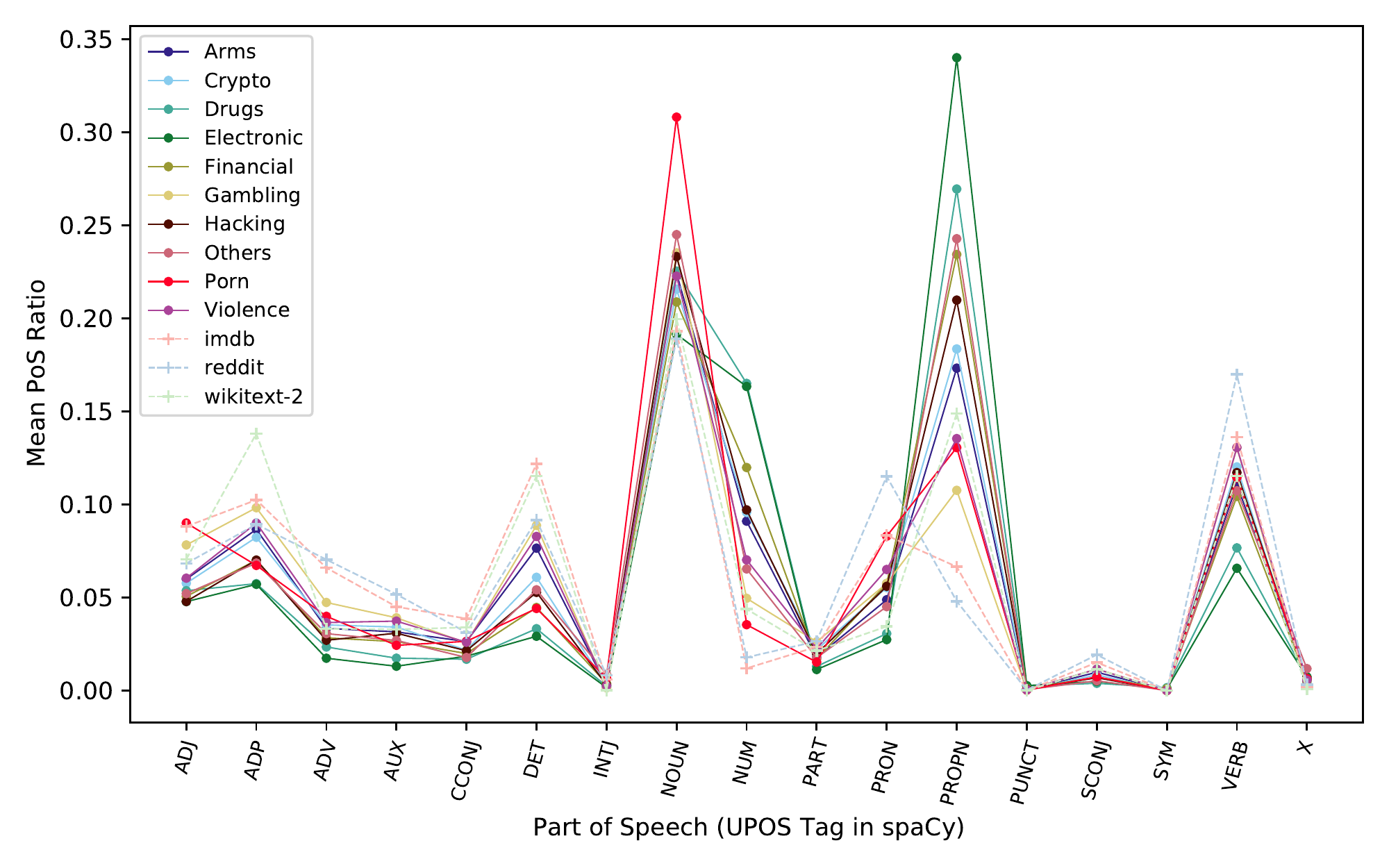}
    \vspace{-20pt}
    \caption{PoS distribution of categories in {\dataset} and Surface Web aggregate data}
    \label{fig:pos_dist}
\end{figure*}

\begin{figure}[ht]
    \centering
    \includegraphics[width=\columnwidth,trim={0.2cm 0.2cm 0.2cm 0.2cm},clip]{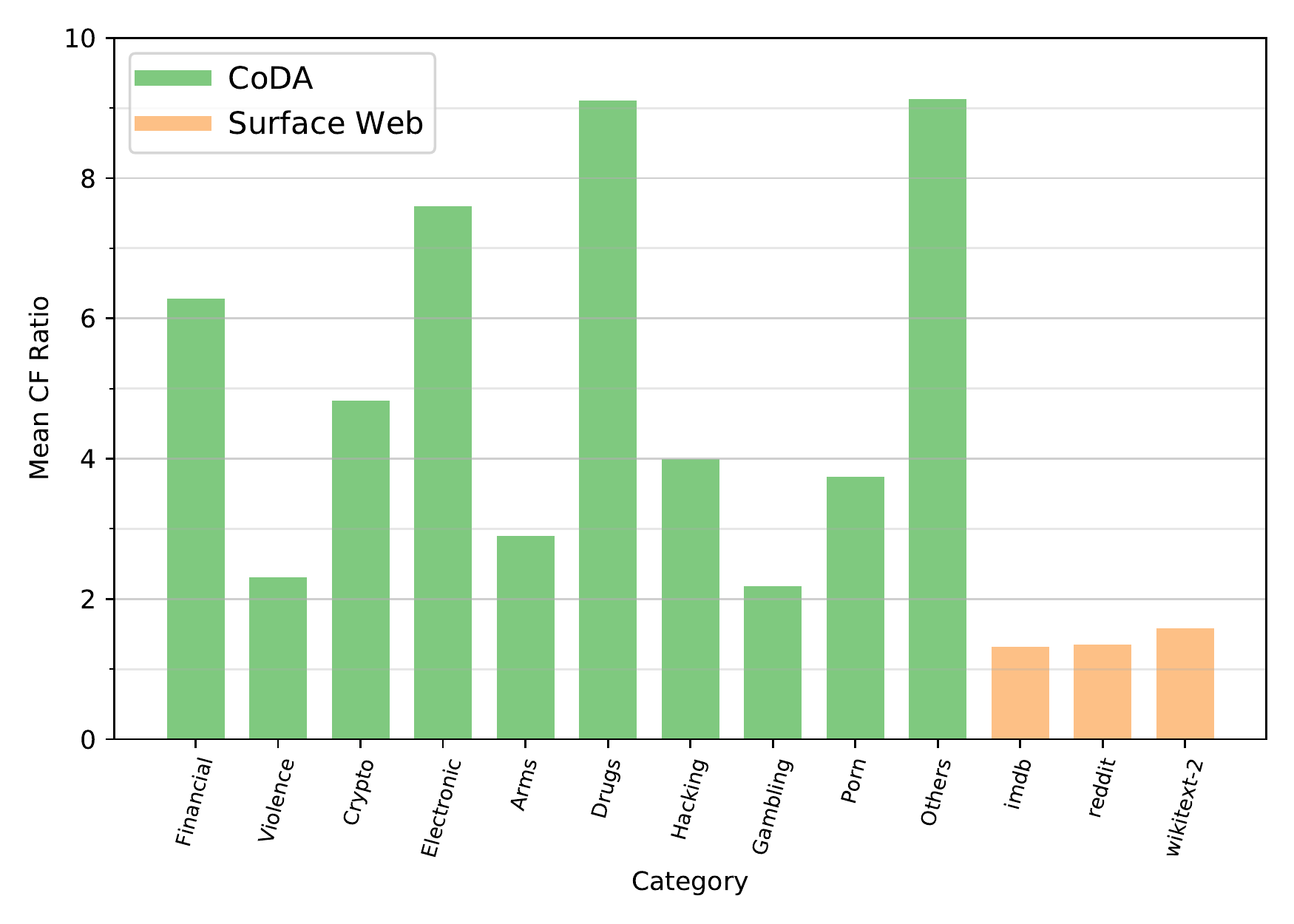}
    \caption{Mean content word / function word (CF) ratio ($\bar{r}_{cf}$) of categories in {\dataset} and Surface Web aggregate data}
    \label{fig:cf_ratio}
\end{figure}

The results for the PoS distribution and CF ratio are shown in Figures~\ref{fig:pos_dist} and \ref{fig:cf_ratio}. It is evident that Dark Web categories generally have a much higher CF ratio compared to that of the Surface Web. From our PoS distribution analysis, we find that the Dark Web documents have a very high ratio of proper nouns (\texttt{PROPN}) and numerals (\texttt{NUM}) compared to the Surface Web documents, both of which are PoS tags of content words. Moreover, the Surface Web documents have a higher ratio of determiners\footnote{Words that modify nouns or noun phrases such as articles} (\texttt{DET}) compared to that of the Dark Web. Since function words serve as critical components of sentence structures, this result implies that language used in the Dark Web may contain a higher proportion of non-sentence structures. For example, texts in the \textit{drugs} category mostly consist of a list of drugs with their price and weight.

\subsection{Discussion}

The use of spaCy for the comparison of PoS distribution and CF ratios may raise questions as spaCy has not been pretrained on Dark Web content, which may yield a higher error rate on the Dark Web results. However, this is not an issue for our work as the main purpose of this analysis is to show that there are linguistic differences between the Dark Web and the Surface Web. If the result of the analysis is heavily skewed by the presence of Dark Web content and additional training is necessary for the pretrained spaCy model, then this implies that there are meaningful lexical differences in the Dark Web. On the other hand, if the result is not particularly affected by the Dark Web content, then it shows that there are clear differences between the two domains. In either case, it is observable that the results suggest the existence of lexical differences between the Dark Web and the Surface Web.

However, there is one possible limitation in our analysis in that the aggregate Surface Web dataset may not encompass the complete representation of the Surface Web. This is evident from observing that a significant portion of the aggregate dataset (IMDb, Wikitext) consists of text content that is comprised of complete sentences. For example, the inclusion of marketplace content such as eBay or Amazon could affect the PoS distribution and the CF ratio of the aggregate Surface Web data. For future work, we may attempt additional analysis through boilerplate removal of noisy, non-structural texts in \dataset{} and the aggregate Surface Web data along with the augmentation of a broader diversity of Surface Web content and examine if this approach significantly affects the results obtained in Section \ref{sec:pos-cf}.

\begin{figure*}[!ht]
    \centering
    \includegraphics[width=\textwidth,trim={0.2cm 0.2cm 0.2cm 0.2cm},clip]{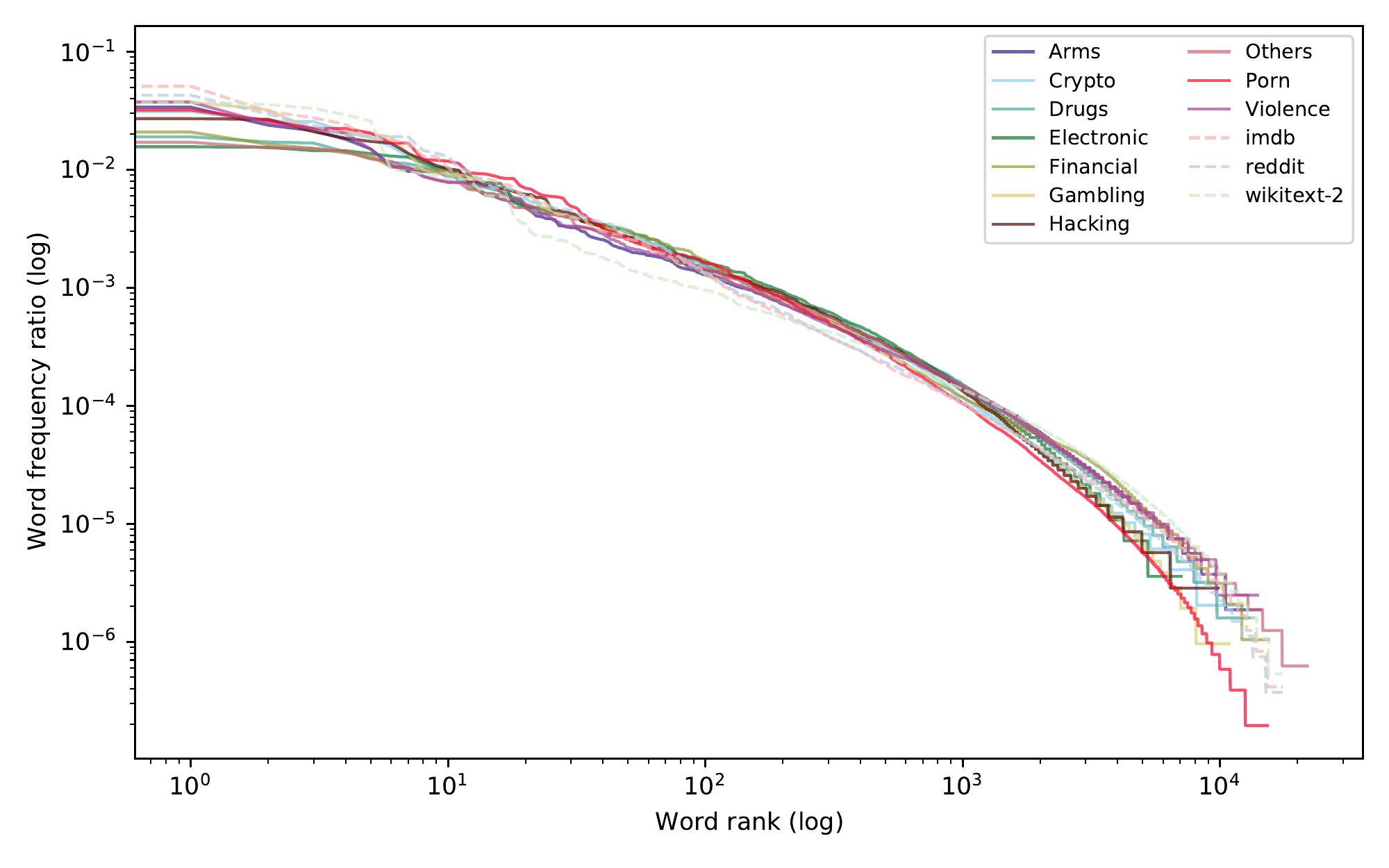}
    \vspace{-20pt}
    \caption{Log-log plot of word frequency distribution of {\dataset} and Surface Web aggregate data by category. We exclude DUTA as it contains too many categories to be presented in the figure.}
    \label{fig:wordfreq_dist}
\end{figure*}

\begin{figure*}[ht]
    \centering
    \includegraphics[width=\textwidth,trim={0.2cm 0.2cm 0.2cm 0.2cm},clip]{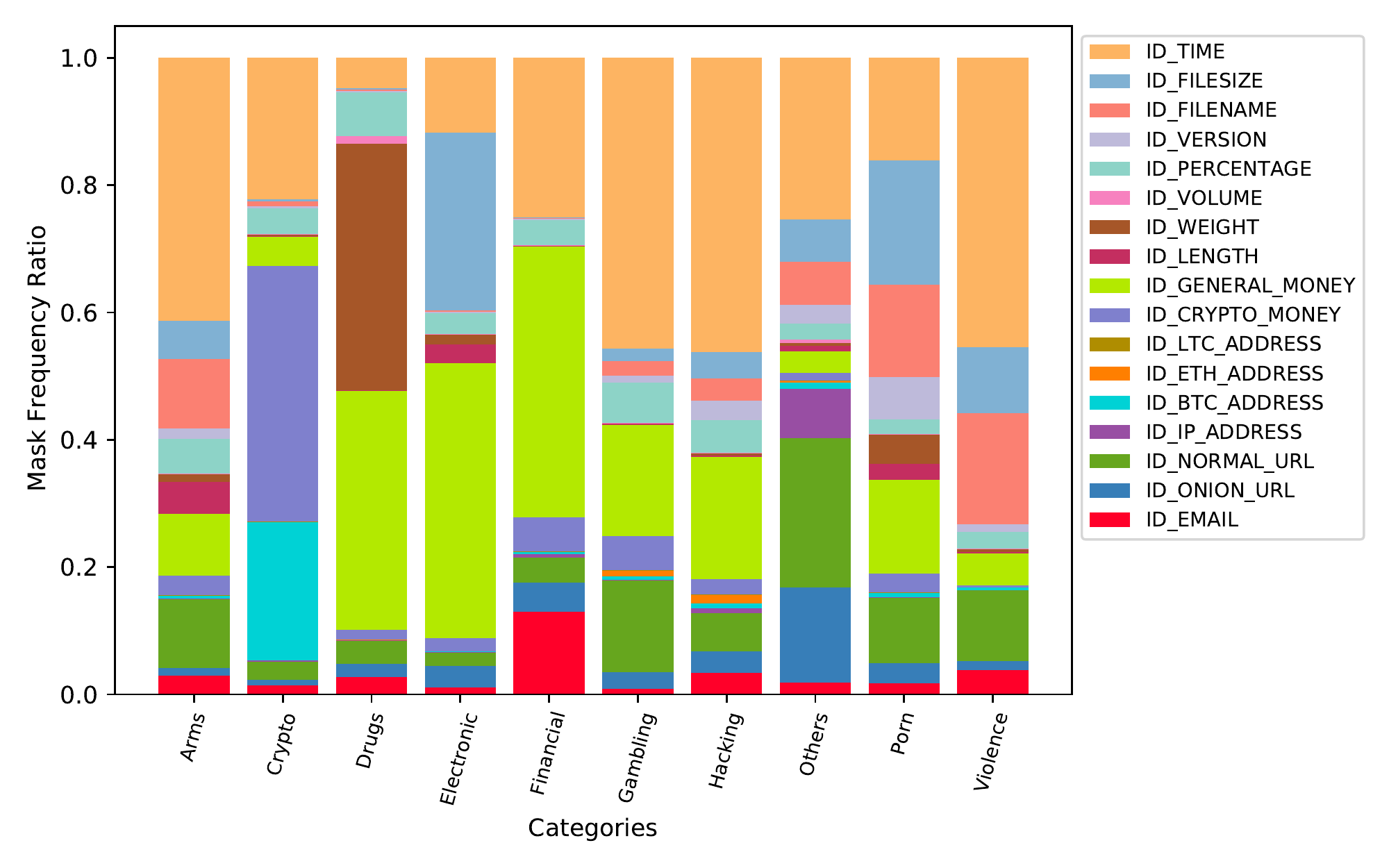}
    \caption{Mask token distribution by category (excluding \texttt{ID\_NUMBER})}
    \label{fig:mask_dist}
\end{figure*}

\clearpage

\section{TF-IDF Measurements}
\label{sec:appendix-tfidf}

A table of TF-IDF measurements (as mentioned in Section~\ref{sec:mask-tfidf}) showing the relevant words and phrases of selected categories in {\dataset} is listed here. 

\begin{table*}[ht]
    \centering
    \resizebox{\textwidth}{!}{%
    \begin{tabular}{c|lr|lr|lr|lr}
\hline
\multicolumn{1}{l|}{} & \multicolumn{2}{c|}{\textbf{Crypto}}   & \multicolumn{2}{c|}{\textbf{Drugs}}   & \multicolumn{2}{c|}{\textbf{Electronics}} & \multicolumn{2}{c}{\textbf{Financial}} \\
\textbf{Rank}         & \textbf{Term}       & \textbf{TF-IDF}  & \textbf{Term}       & \textbf{TF-IDF} & \textbf{Term}         & \textbf{TF-IDF}   & \textbf{Term}       & \textbf{TF-IDF}  \\ \hline
1                     & \texttt{ID\_NUMBER}          & 0.684            & \texttt{ID\_NUMBER}          & 0.863           & \texttt{ID\_NUMBER}            & 0.799             & \texttt{ID\_NUMBER}          & 0.906            \\
2                     & \texttt{ID\_CRYPTO\_MONEY}   & 0.418            & \texttt{ID\_WEIGHT}          & 0.255           & \texttt{ID\_GENERAL\_MONEY}    & 0.291             & \texttt{ID\_GENERAL\_MONEY}  & 0.220            \\
3                     & bitcoin             & 0.354            & \texttt{ID\_GENERAL\_MONEY}  & 0.249           & iphone                & 0.279             & card                & 0.215            \\
4                     & \texttt{ID\_TIME}            & 0.233            & buy                 & 0.112           & \texttt{ID\_FILESIZE}          & 0.184             & \texttt{ID\_TIME}            & 0.130            \\
5                     & \texttt{ID\_BTC\_ADDRESS}    & 0.227            & weed                & 0.101           & apple                 & 0.142             & \texttt{ID\_EMAIL}           & 0.067            \\
6                     & btc                 & 0.110            & pot                 & 0.089           & pro                   & 0.111             & buy                 & 0.065            \\
7                     & use                 & 0.071            & pill                & 0.084           & ipad                  & 0.110             & account             & 0.051            \\
8                     & wallet              & 0.069            & online              & 0.080           & macbook               & 0.092             & credit              & 0.048            \\
9                     & address             & 0.068            & cocaine             & 0.073           & imac                  & 0.089             & paypal              & 0.048            \\
10                    & buy                 & 0.068            & cannabis            & 0.068           & airpod                & 0.082             & transfer            & 0.044            \\
11                    & transaction         & 0.061            & lsd                 & 0.057           & \texttt{ID\_TIME}              & 0.079             & order               & 0.039            \\
12                    & get                 & 0.060            & mdma                & 0.051           & buy                   & 0.073             & get                 & 0.038            \\
13                    & blockchain          & 0.057            & drug                & 0.050           & card                  & 0.071             & dump                & 0.038            \\
14                    & invest              & 0.049            & adderall            & 0.050           & watch                 & 0.058             & cc                  & 0.037            \\
15                    & \texttt{ID\_GENERAL\_MONEY}  & 0.048            & viagra              & 0.049           & ipod                  & 0.055             & good                & 0.036            \\
16                    & coin                & 0.046            & xanax               & 0.049           & case                  & 0.052             & money               & 0.036            \\
17                    & service             & 0.046            & \texttt{ID\_PERCENTAGE}      & 0.046           & gopro                 & 0.051             & new                 & 0.036            \\
18                    & make                & 0.045            & product             & 0.045           & product               & 0.051             & use                 & 0.035            \\
19                    & \texttt{ID\_PERCENTAGE}      & 0.042            & order               & 0.044           & xs                    & 0.051             & cvv                 & 0.034            \\
20                    & double              & 0.041            & quality             & 0.042           & order                 & 0.047             & shop                & 0.034            \\ \hline
\multicolumn{1}{l|}{} & \multicolumn{2}{c|}{\textbf{Gambling}} & \multicolumn{2}{c|}{\textbf{Hacking}} & \multicolumn{2}{c|}{\textbf{Pornography}} & \multicolumn{2}{c}{\textbf{Violence}}  \\
\textbf{Rank}         & \textbf{Term}       & \textbf{TF-IDF}  & \textbf{Term}       & \textbf{TF-IDF} & \textbf{Term}         & \textbf{TF-IDF}   & \textbf{Term}       & \textbf{TF-IDF}  \\ \hline
1                     & \texttt{ID\_NUMBER}          & 0.593            & \texttt{ID\_NUMBER}          & 0.792           & porno                 & 0.605             & \texttt{ID\_NUMBER}          & 0.915            \\
2                     & casino              & 0.396            & hack                & 0.364           & porn                  & 0.557             & \texttt{ID\_TIME}            & 0.173            \\
3                     & game                & 0.256            & facebook            & 0.262           & video                 & 0.272             & kill                & 0.074            \\
4                     & br                  & 0.190            & \texttt{ID\_TIME}            & 0.167           & free                  & 0.220             & anonymous           & 0.066            \\
5                     & online              & 0.177            & account             & 0.162           & sex                   & 0.154             & hitman              & 0.066            \\
6                     & slot                & 0.173            & hacker              & 0.092           & girl                  & 0.125             & \texttt{ID\_FILENAME}        & 0.065            \\
7                     & play                & 0.161            & password            & 0.087           & teen                  & 0.123             & murder              & 0.064            \\
8                     & \texttt{ID\_TIME}            & 0.103            & use                 & 0.070           & \texttt{ID\_NUMBER}            & 0.105             & people              & 0.063            \\
9                     & bet                 & 0.098            & \texttt{ID\_GENERAL\_MONEY}  & 0.068           & boy                   & 0.093             & like                & 0.056            \\
10                    & win                 & 0.096            & service             & 0.059           & fuck                  & 0.091             & one                 & 0.054            \\
11                    & poker               & 0.094            & email               & 0.058           & child                 & 0.083             & get                 & 0.052            \\
12                    & page                & 0.094            & software            & 0.052           & cock                  & 0.077             & file                & 0.052            \\
13                    & get                 & 0.093            & ransomware          & 0.050           & cp                    & 0.070             & post                & 0.047            \\
14                    & free                & 0.089            & download            & 0.045           & young                 & 0.068             & site                & 0.043            \\
15                    & money               & 0.087            & get                 & 0.044           & pussy                 & 0.053             & comment             & 0.043            \\
16                    & time                & 0.087            & free                & 0.040           & pedo                  & 0.053             & \texttt{ID\_NORMAL\_URL}     & 0.042            \\
17                    & card                & 0.083            & instagram           & 0.039           & say                   & 0.051             & hire                & 0.042            \\
18                    & player              & 0.081            & attack              & 0.039           & mom                   & 0.050             & make                & 0.041            \\
19                    & good                & 0.077            & hacking             & 0.039           & gay                   & 0.050             & say                 & 0.040            \\
20                    & roulette            & 0.075            & online              & 0.038           & get                   & 0.049             & use                 & 0.040            \\ \hline
\end{tabular}%
    }
    \caption{Terms with the highest TF-IDF for selected categories in {\dataset}}
    \label{tab:tfidf}
\end{table*}

\newpage

\section{Forum \& Marketplace Benchmark Dataset}
\label{sec:appendix-usecase2}

\begin{table*}[h]
    \centering
    \resizebox{\textwidth}{!}{%
    \begin{tabular}{@{}llll@{}}
        \toprule
        \textbf{Category} & \textbf{Website title} & \textbf{Onion URL} & \textbf{\# of pages} \\ 
        \midrule
        Drugs   &  Ang*******            &  ang*****************************************************.onion   &  377     \\
        Drugs   &  Glo**********         &  ny4*************.onion   &  307     \\
        Drugs   &  Opi***********        &  opi*************.onion   &  215     \\
        Drugs   &  Pot****               &  pot*****************************************************.onion   &  19     \\
        Drugs   &  Eu*****               &  wge*****************************************************.onion   &  13     \\
        Drugs   &  Kam****************   &  bep*****************************************************.onion   &  5     \\
        \midrule
        Financial   &  Wal********       &  z2h*************.onion   &  241     \\
        Financial   &  Tor*****          &  tor*****************************************************.onion   &  71     \\
        Financial   &  Cov*********      &  cov*****************************************************.onion   &  62     \\
        Financial   &  Fin************   &  fin*************.onion   &  51     \\
        Financial   &  Cas*****          &  hss*****************************************************.onion   &  35     \\
        Financial   &  Car******         &  car*************.onion   &  25     \\
        Financial   &  Imp***********    &  srw*****************************************************.onion   &  16     \\
        Financial   &  Lig********       &  sw3*****************************************************.onion   &  14     \\
        Financial   &  Kin*********      &  kin*****************************************************.onion   &  13     \\
        Financial   &  Cou******************   &  cou*************.onion   &  12     \\
        Financial   &  Cas**********     &  maf*****************************************************.onion   &  11     \\
        Financial   &  Kry*********      &  kry*************.onion   &  10     \\
        Financial   &  The*************  &  nar*************.onion   &  9     \\
        Financial   &  Pre**********     &  hbl*****************************************************.onion   &  8     \\
        Financial   &  Hor**********     &  hor*************.onion   &  8     \\
        Financial   &  Ban***            &  ban*************.onion   &  7     \\
        Financial   &  Tor************** &  vrm*************.onion   &  7     \\
        Financial   &  net****           &  net*************.onion   &  6     \\
        Financial   &  Bla*********      &  bla*************.onion   &  6     \\
        Financial   &  LAR***********    &  fiw*****************************************************.onion   &  5     \\
        Financial   &  eas********       &  eas*************.onion   &  3     \\
        Financial   &  CHE********       &  o6k*************.onion   &  3     \\
        Financial   &  PAY*********      &  ity*************.onion   &  3     \\
        \midrule
        Weapons   &  Glo********         &  glo*************.onion   &  266   \\
        Weapons   &  Alp********         &  alp*************.onion   &  219   \\
        Weapons   &  Exe***************  &  5zk*************.onion   &  181   \\
        Weapons   &  Eur******           &  hyj*****************************************************.onion   &  4     \\
        Weapons   &  UK********************   &  tuu*************.onion   &  4     \\
        \midrule
        \textbf{Total Drugs}        &     &    &  936    \\
        \textbf{Total Finance}      &     &    &  626    \\
        \textbf{Total Weapons}      &     &    &  674    \\
        \textbf{All}                &     &    &  \textbf{2236} \\
        \bottomrule
    \end{tabular}%
    }
    \caption{Source of our forum \& marketplace benchmark dataset as described in Section~\ref{sec:usecase}. To follow ethical guidelines, we mask the website titles and onion addresses.}
    \label{tab:usecase2-data}
\end{table*}

\end{document}